\documentclass{article}
\PassOptionsToPackage{table}{xcolor}
\usepackage[T1]{fontenc}
\usepackage[preprint]{colm2026_conference}
\usepackage{latexsym}

\usepackage[utf8]{inputenc}

\usepackage{microtype}

\usepackage{inconsolata}

\usepackage{graphicx}
\usepackage{algorithm}
\usepackage{algpseudocode}
\usepackage{textcomp}
\usepackage{newfloat}
\usepackage{listings}
\usepackage{booktabs}
\usepackage{amsmath}
\usepackage{amsfonts}
\usepackage{xcolor} 
\usepackage{multirow}
\usepackage{makecell}
\usepackage{subcaption}
\usepackage{amssymb}
\usepackage{pifont}
\usepackage{tikz}
\usepackage[normalem]{ulem}
\usepackage{mathtools}

\usepackage{soul} 

\definecolor{BenchGreen}{RGB}{3,151,92}
\definecolor{GainGreen}{RGB}{0,100,60}
\definecolor{BenchRed}{RGB}{201,52,52}
\definecolor{BenchOrange}{RGB}{255,129,3}
\definecolor{ResultBest}{RGB}{255,239,194}
\newcommand{\bestscore}[1]{\cellcolor{ResultBest}\textbf{#1}}
\newcommand{\secondscore}[1]{\cellcolor{blue!5}\underline{#1}}
\definecolor{ResultThird}{RGB}{228,241,230}
\newcommand{\thirdscore}[1]{\cellcolor{ResultThird}#1}
\DeclareRobustCommand{\deepseekicon}{\raisebox{-0.18\height}{\includegraphics[height=2.13ex]{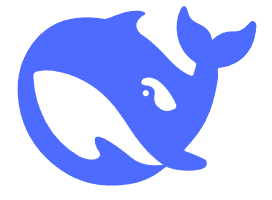}}}
\DeclareRobustCommand{\glmicon}{\raisebox{-0.20\height}{\includegraphics[height=2.20ex]{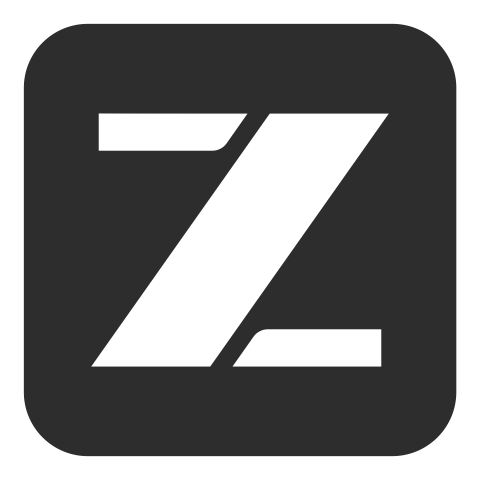}}}
\DeclareRobustCommand{\qwenicon}{\raisebox{-0.3\height}{\includegraphics[height=3.5ex]{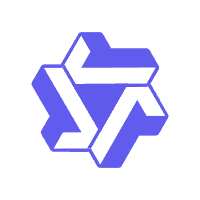}}}
\DeclareRobustCommand{\chatgpticon}{\raisebox{-0.20\height}{\includegraphics[height=2.20ex]{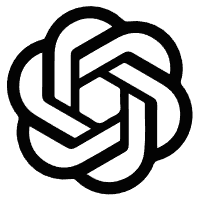}}}
\DeclareRobustCommand{\trainmark}{\tikz[baseline=-0.1ex,x=0.85ex,y=0.85ex]{%
  \fill[BenchOrange] (0.5,0) .. controls (-0.05,0) and (-0.12,0.55) .. (0.22,0.85)
    .. controls (0.22,0.62) and (0.40,0.60) .. (0.40,0.82)
    .. controls (0.39,1.04) and (0.63,1.20) .. (0.55,1.50)
    .. controls (1.16,1.00) and (1.20,0) .. (0.5,0) -- cycle;
  \fill[yellow!40] (0.5,0.12) .. controls (0.22,0.15) and (0.38,0.53) .. (0.58,0.76)
    .. controls (0.53,0.43) and (0.94,0.17) .. (0.5,0.12) -- cycle;}}

\DeclareRobustCommand{\freezemark}{\tikz[baseline=-0.65ex,x=1ex,y=1ex]{%
  \foreach \angle in {0,60,...,300}{
    \draw[cyan!45!gray!60!white,line width=0.35pt,line cap=round,rotate=\angle]
      (0,0) -- (0,0.65) (0,0.38) -- (-0.18,0.53)
      (0,0.38) -- (0.18,0.53);
  }}}

\newcommand{\gain}[1]{\textcolor{GainGreen}{\raisebox{-0.45ex}{\scalebox{0.72}{$\boldsymbol{\uparrow}$\textbf{#1}}}}}

\usepackage{wrapfig}
\usepackage{hyperref}
\hypersetup{colorlinks=true,linkcolor=blue!50!black,
  citecolor=blue!50!green,urlcolor=blue!60!black}

\let\cite\citep
\usepackage{tabularx}
\usepackage{fontawesome5}
\hypersetup{
  colorlinks=true,
  citecolor=blue!50!green,
  urlcolor=blue!50!green
}
\hypersetup{
  colorlinks=true,
  linkcolor=red
}
\usepackage{adjustbox}

\newcommand{\paperrunningtitle}{AtomSD: Procedural Atom-Guided Self-Distillation for LLM Agents}
\title{%
  \makebox[\textwidth][c]{%
    \parbox{1.2\textwidth}{%
      \centering
      \textsc{AtomSD}: Procedural Atom-Guided Self-Distillation for \\LLM Agents
    }%
  }%
}
\author{%
  \makebox[\dimexpr\textwidth-2\tabcolsep\relax][c]{%
  \parbox{1.10\textwidth}{%
    \centering
    Bohan Yu$^{1,2}$,
    Pengfei Cao$^{2,3}$\setcounter{footnote}{1}\thanks{\,\,Corresponding authors.},
    Kai Tian$^{4}$,
    Chen Han$^{1}$,
    Shi-Yang Li$^{1,2}$, 
    Chenxi Zhou$^{1,2}$, \\
    Junlin Liu$^{2,3}$,
    Cao Liu$^{4}$,
    Ke Zeng$^{4}$,
    Jun Zhao$^{2,3}$,
    Kang Liu$^{2,3}$\footnotemark[2] \\
    \normalfont
    $^1$School of Advanced Interdisciplinary Sciences, University of Chinese Academy of Sciences \\
    $^2$The Key Laboratory of Cognition and Decision Intelligence for Complex Systems,\\ Institute of Automation, Chinese Academy of Sciences \\
    $^3$School of Artificial Intelligence, University of Chinese Academy of Sciences \\
    $^4$Meituan \\
    \texttt{yubohan2025@ia.ac.cn \quad \{pengfei.cao,jzhao,kliu\}@nlpr.ia.ac.cn}
    }%
  }%
}
\begin{document}
\maketitle
\pagestyle{fancy}
\fancyhead{}
\fancyhead[L]{\paperrunningtitle}

\fancypagestyle{firstpage}{%
  \fancyhead{}
  \fancyhead[L]{%
    \includegraphics[height=0.65cm]{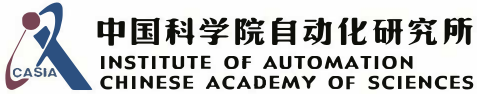}\hspace{0.5cm}%
    \includegraphics[height=0.65cm]{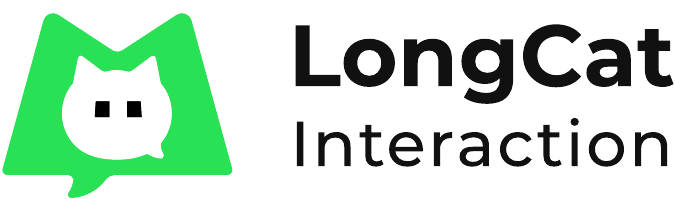}}
}
\thispagestyle{firstpage}

\begin{abstract}
Large language models (LLMs) possess broad factual knowledge and strong reasoning capabilities, but solving complex tasks also requires procedural knowledge about how to act.
Existing approaches often represent procedural knowledge through trajectories or bundled skills, limiting flexible reuse across interaction histories. This paper introduces \textbf{Procedural Atoms}, fine-grained and independently applicable units that represent \emph{Intent}, \emph{When}, \emph{Do}, and \emph{Avoid} guidance.
Building on this representation, we propose \textbf{AtomSD}, an \textbf{Atom}-guided \textbf{S}elf-\textbf{D}istillation framework in which the learner continually extracts Procedural Atoms from its own successful trajectories and distills them into the evolving policy through Atom-conditioned supervision and task-reward optimization.
Comprehensive experiments show that AtomSD consistently outperforms strong reward-based and distillation methods, exceeding the state-of-the-art baseline by up to 9.31 points in average performance.
Further analyses demonstrate that the proposed Procedural Atoms provide effective guidance, highlighting the value of atom-level procedural knowledge for policy learning. 
\href{https://github.com/SharkSpicy-NLP/AtomSD}{%
  \faGithub\enspace\textbf{GitHub}%
}
\end{abstract}

\section{Introduction}
Large language models (LLMs) have acquired extensive factual knowledge and strong reasoning capabilities, enabling them to tackle increasingly complex tasks across diverse domains~\cite{openai2024gpt4technicalreport,zhao2023survey}. However, knowing the relevant facts does not necessarily imply knowing how to act. 
For example, \textit{a model may understand the properties of every reagent in a scientific experiment, yet still fail to execute the experimental procedure correctly.}
Such failures reflect the difference between declarative knowledge about a domain and procedural knowledge or planning knowledge about how tasks within that domain should be carried out.
It becomes particularly consequential when successful behavior depends on sequences of interdependent decisions rather than isolated answers. 
In such contexts, an agent must determine which operations advance a goal, when their prerequisites hold, how intermediate results constrain later actions, and how its procedure should adapt as conditions change~\cite{shridhar2021alfworldaligningtextembodied,wang2022scienceworldagentsmarter5th,yao2023webshopscalablerealworldweb,yu2026ruleweaverbenchmarkingrulecenteredscenario}. Therefore, learning reusable procedural knowledge is an important step toward translating the broad capabilities of LLMs into reliable performance in complex tasks.

Toward this goal, existing approaches leverage interaction trajectories to construct and transfer reusable procedural knowledge, broadly following two directions.
\noindent\textbf{(1) \emph{Organizing trajectories into reusable guidance}.} Prior work transforms previous interactions into reflections, workflows, skills, or conditional rules that can guide subsequent behavior through prompting~\cite{Agent_Workflow_Memory,mi2026skillpro,liang-etal-2026-learning-remember,sun-etal-2026-shuttle-between-symbolic-instruction}. Such representations abstract beyond individual actions and make past experience accessible across future interactions.
\noindent\textbf{(2) \emph{Distilling experience into the policy}.} A second line of work transfers knowledge from demonstrations, feedback, or skill-augmented teacher context into the learner itself, allowing the resulting policy to benefit from this supervision without requiring the same external guidance at inference time~\cite{opsd,sdpo,skill_sd,sdar}.

\begin{wrapfigure}{r}{0.55\textwidth}
    \centering
    \includegraphics[width=\linewidth]{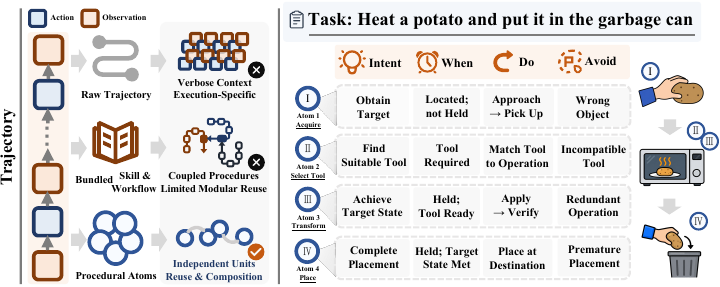}
    \caption{Procedural Atoms enable fine-grained reuse and composition of procedural knowledge.}
    \label{fig:procedural-knowledge}
\end{wrapfigure}
Despite their effectiveness, these approaches leave two important questions unresolved.
\noindent(1) \emph{What should be the reusable unit of procedural knowledge?} As shown in Figure~\ref{fig:procedural-knowledge}, complete trajectories retain instance-specific objects, observations, and action sequences, while workflows and skills often bundle multiple operations into a single unit of guidance. Both can limit transfer across interaction histories: a new execution may require only a subset of the operations in a previous procedure, or may need to compose operations learned from different experiences.
This calls for finer-grained procedural units that can be independently reused and flexibly composed across interaction histories.
\noindent(2) \emph{How should procedural knowledge evolve with the learner?}
Existing approaches often rely on separately maintained skill libraries or external models to derive and update guidance from interaction trajectories~\cite{mi2026skillpro,sdar,skill_sd,yang2026opidonpolicyskilldistillation}.
However, reusable procedural knowledge should ideally co-evolve with the policy itself: as the learner acquires new capabilities and succeeds on new interaction histories, the knowledge available for subsequent learning should be updated accordingly.
In a word, the aforementioned limitations motivate us to ask: \textbf{\emph{what form of procedural knowledge can be extracted from previous interactions, and continuously distilled as the learner evolves?}}

To this end, this paper proposes \textbf{AtomSD}, a procedural \textbf{\uwave{Atom}}-guided \textbf{\uwave{S}}elf-\textbf{\uwave{D}}istillation framework that couples continual procedural abstraction with online policy learning. 
To provide a reusable unit of procedural knowledge, we define a \emph{Procedural Atom} as a fine-grained and independently applicable unit of procedural knowledge. As illustrated in Figure~\ref{fig:procedural-knowledge}, a Procedural Atom consists of (i) \emph{Intent}, specifying what the procedure is intended to accomplish; (ii) \emph{When}, defining its applicability conditions; (iii) \emph{Do}, specifying an ordered procedure; and (iv) \emph{Avoid}, identifying operations or failure modes to avoid.
With this goal-directed and condition-aware representation, individual Atoms can be selectively applied and flexibly composed across different interaction histories, rather than reused only in complete and inseparable trajectories or bundled workflows.
In this way, procedural knowledge can better adapt to new interactions.

To enable procedural knowledge to co-evolve with the learner, AtomSD establishes a self-improving loop through two complementary processes.
\noindent\textbf{(1) \emph{Online procedural abstraction}.}
At each training iteration, the student acts as its own knowledge extractor, abstracting successful on-policy trajectories into independently applicable Procedural Atoms. 
By sharing the resulting Atom set within the same task, knowledge extracted from one trajectory can guide other interaction histories, including both successful and failed ones.
As the policy evolves, subsequent extraction continually refreshes the procedural guidance to reflect the learner's growing capabilities.
\noindent\textbf{(2) \emph{Atom-guided self-distillation}.}
An Atom-conditioned teacher, derived from the student, evaluates the original on-policy trajectories, transforming self-extracted procedural knowledge into token-level supervision across different interaction histories. Through joint reward optimization and self-distillation, this guidance is progressively internalized into the policy, whose continued improvement in turn supports further procedural knowledge discovery.

Comprehensive experiments on widely used benchmarks~\cite{shridhar2021alfworldaligningtextembodied,wang2022scienceworldagentsmarter5th,yao2023webshopscalablerealworldweb,guo2025stabletoolbenchstablelargescalebenchmarking} demonstrate that AtomSD consistently outperforms six strong reward-based and distillation baselines, while also showing competitive performance against advanced API-based models such as DeepSeek-V4-Pro~\cite{deepseek2026v4pro} and GPT-5.5~\cite{openai2026gpt55}. AtomSD exceeds the state-of-the-art baseline in average performance by up to 9.31 points, with improvements extending to unseen tasks. Further analyses show that fine-grained Procedural Atoms provide compact and effective guidance, validating the design of Atom-guided self-distillation.

Our key contributions are as follows:
\begin{itemize}
    \item This paper introduces \textbf{Procedural Atoms}, fine-grained and independently applicable units that support selective reuse and flexible composition of procedural knowledge across interaction histories.
    \item \textbf{AtomSD} combines online procedural abstraction with Atom-guided self-distillation, allowing procedural knowledge to be progressively internalized into the policy while co-evolving with the learner.
\end{itemize}

\section{Procedural Atom-Guided Self-Distillation}

\begin{figure*}[htbp]
    \centerline{\includegraphics[scale=0.9]{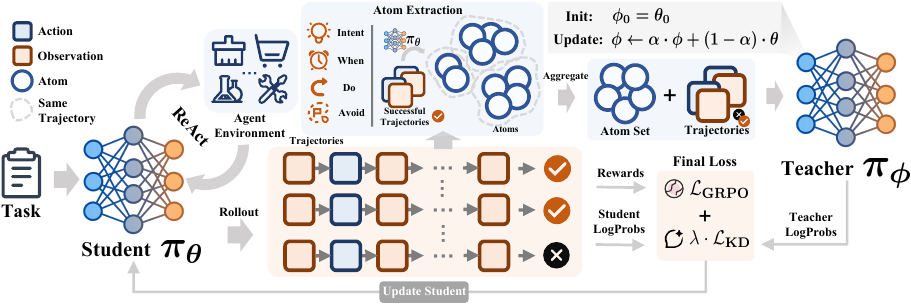}}%
    \caption{Overview of AtomSD. Procedural Atoms are continually abstracted from the student's own successful trajectories and distilled back into the evolving policy through Atom-guided self-distillation.}
    \label{fig:main}
\end{figure*}

As shown in Figure~\ref{fig:main}, this paper proposes \emph{Procedural Atom-Guided Self-Distillation} as a self-improving loop in which procedural knowledge is continually extracted from on-policy interactions and internalized into the evolving student policy.
We define \emph{Procedural Atoms} as fine-grained functional units of procedural knowledge that capture guidance across four aspects: \emph{Intent}, \emph{When}, \emph{Do}, and \emph{Avoid}.
As training proceeds iteratively, the student policy is updated at each training iteration based on its newly generated on-policy interactions. We formulate each training iteration as one complete self-improving loop consisting of two coupled processes: (1) \emph{Online Procedural Abstraction} and (2) \emph{Atom-Guided Self-Distillation}.

\subsection{Online Procedural Abstraction}
\label{sec:online-procedural-abstraction}

We represent an interaction trajectory $\tau$ as a task together with the sequence of observations and actions encountered during execution. From its own successful trajectories, the evolving student serves as the procedural knowledge extractor, abstracting reusable procedures into independently applicable Procedural Atoms.

\paragraph{Atom Structure}

Formally, an Atom $\omega$ is represented as
\begin{equation}
    \omega =
    \left\langle
    \iota_\omega,\mathcal{C}_\omega,
    \mathcal{P}_\omega,\mathcal{B}_\omega
    \right\rangle,
    \label{eq:atom}
\end{equation}
where $\iota_\omega$ denotes the procedural intent;
$\mathcal{C}_\omega=\{c_{\omega,1},\ldots,c_{\omega,n_C}\}$ specifies the conditions under which the Atom applies;
$\mathcal{P}_\omega=(p_{\omega,1},\ldots,p_{\omega,n_P})$ defines the ordered procedure to execute;
and $\mathcal{B}_\omega=\{b_{\omega,1},\ldots,b_{\omega,n_B}\}$ captures optional failure modes or inappropriate operations to avoid.
Here, the elements of $\mathcal{C}_\omega$, $\mathcal{P}_\omega$, and $\mathcal{B}_\omega$ denote the conditions, operations, and avoidance items, respectively, while $n_C$, $n_P$, and $n_B$ denote their respective counts.
Together, these components characterize what to achieve, when to apply the Atom, how to act, and what boundaries to respect.


\begin{wrapfigure}{r}{0.5\textwidth}
\vspace{-0.15em}
\noindent\textbf{Atom Example}\quad
\smallskip
\noindent
The following example illustrates how a Procedural Atom $\omega$ instantiates the four components defined above.
\par\medskip

\noindent
\fbox{%
\begin{minipage}{\dimexpr\linewidth-2\fboxsep-2\fboxrule\relax}
\small
\setlength{\parindent}{0pt}
\setlength{\parskip}{6pt}

\textbf{Intent} $(\iota_\omega)$:
\textit{Clean a held object using a designated
cleaning facility.}

\textbf{When} $(\mathcal{C}_\omega)$:
\textit{The agent holds an object requiring cleaning;
a suitable cleaning facility is available.}

\textbf{Do} $(\mathcal{P}_\omega)$:
\textit{Navigate to the cleaning facility if not
already there; execute the cleaning action with
the held object and the facility.}

\textbf{Avoid} $(\mathcal{B}_\omega)$:
\textit{Attempting to clean without reaching the
facility; cleaning an object that is not currently
held; using an incorrect or unavailable cleaning
method.}

\end{minipage}%
}
\end{wrapfigure}

\paragraph{Atomization and Sharing}
To provide fine-grained procedural knowledge for reuse across interaction histories, we atomize successful trajectories and aggregate their Atoms at the task level.
Consider the $k$-th self-improving iteration and a sampled task $x$ in its training batch. Let $\theta_k$ denote the student parameters at the beginning of iteration $k$. For task $x$, the current student $\pi_{\theta_k}$ generates $G$ on-policy trajectories $\mathcal{T}_x^{(k)}=\{\tau_i\}_{i=1}^{G}$ without access to Procedural Atoms.
Within $\mathcal{T}_x^{(k)}$, we identify the successful trajectories as
\begin{equation}
    \mathcal{S}_x^{(k)}
    =
    \{\tau_i\in\mathcal{T}_x^{(k)}\mid s(\tau_i)=1\},
    \label{eq:successful-trajectories}
\end{equation}
where $s(\tau_i)\in\{0,1\}$ indicates whether $\tau_i$ successfully completes task $x$.
The current student then serves as its own procedural knowledge extractor, atomizing each successful trajectory $\tau_i\in\mathcal{S}_x^{(k)}$ using the extraction prompt shown in Figure~\ref{fig:prompt_extract}:
\begin{equation}
\begin{aligned}
    \mathcal{A}_i^{(k)}
    &=
    \textproc{Extract}\bigl(\pi_{\theta_k};\tau_i\bigr)
    =
    \left\{
    \omega_{i,r}
    =
    \left\langle
    \iota_{\omega_{i,r}},
    \mathcal{C}_{\omega_{i,r}},
    \mathcal{P}_{\omega_{i,r}},
    \mathcal{B}_{\omega_{i,r}}
    \right\rangle
    \right\}_{r=1}^{M_i},
\end{aligned}
\label{eq:trajectory-atomization}
\end{equation}
where $M_i$ denotes the number of Procedural Atoms extracted from $\tau_i$. The trajectory-level Atom sets are then aggregated into the task-level procedural knowledge of the current self-improving loop:
\begin{equation}
    \Omega_x^{(k)}
    =
    \bigcup_{\tau_i\in\mathcal{S}_x^{(k)}}\mathcal{A}_i^{(k)}.
    \label{eq:atom-extraction}
\end{equation}

\begin{wrapfigure}[21]{r}{0.5\textwidth}
\vspace{-1.5em}
\begin{minipage}{\linewidth}
\begin{algorithm}[H]
  \caption{AtomSD: Procedural Atom-Guided Self-Distillation}
  \label{alg:atomsd}
  \scriptsize
  \begin{algorithmic}[1]
    \Require Initial student policy $\pi_{\theta_0}$; task distribution $\mathcal{D}$; number of on-policy trajectories per task $G$; distillation weight $\lambda$; clipping range $\epsilon$; EMA decay $\alpha$; training iterations $K$
    \Ensure Trained student policy $\pi_{\theta_K}$

    \State Initialize teacher parameters $\phi_0\gets\theta_0$
    \For{$k\gets0$ \textbf{to} $K-1$}
      \State Sample a training batch $\mathcal{X}^{(k)}\sim\mathcal{D}$

      \ForAll{$x\in\mathcal{X}^{(k)}$}
      \Statex {\setlength{\fboxsep}{1pt}%
\colorbox{blue!8}{\textbf{(1) Online Procedural Abstraction}}}
       \State Generate $\mathcal{T}_x^{(k)}=\{\tau_i\sim\pi_{\theta_k}(\cdot\mid x)\}_{i=1}^{G}$
        \State $\mathcal{S}_x^{(k)}\gets\{\tau_i\in\mathcal{T}_x^{(k)}\mid s(\tau_i)=1\}$
        \Comment{successful trajectories}
        \State $\mathcal{A}_i^{(k)}\gets\textproc{Extract}(\pi_{\theta_k};\tau_i)=\{\omega_{i,r}\}_{r=1}^{M_i}$, $\forall\tau_i\in\mathcal{S}_x^{(k)}$ 
        \State $\Omega_x^{(k)}\gets\displaystyle\bigcup_{\tau_i\in\mathcal{S}_x^{(k)}}\mathcal{A}_i^{(k)}$
        \Comment{task-level Atom set}

        \Statex {\setlength{\fboxsep}{1pt}%
\colorbox{blue!8}{\textbf{(2) Atom-Guided Self-Distillation}}}
        \State Compute $A_i^{\mathrm{RL},(k)}$ for all $\tau_i\in\mathcal{T}_x^{(k)}$
        \Comment{reward signal}
        \State $q_{ij}^{\mathrm{Atom},(k)}
        \gets\pi_{\phi_k}(y_{ij}\mid h_{ij},\Omega_x^{(k)})$
        \Comment{Atom support}
        \State Compute $A_{ij}^{\mathrm{Atom},(k)}$ from $q_{ij}^{\mathrm{Atom},(k)}$ using Eq.~(\ref{eq:atom-advantage})
      \EndFor
        \State $\theta_{k+1}\gets\textproc{Update}(\theta_k)$ using all $A_i^{\mathrm{RL},(k)}$ and $A_{ij}^{\mathrm{Atom},(k)}$ over $\mathcal{X}^{(k)}$ via Eq.~(\ref{eq:joint-loss}) with $\lambda$ and $\epsilon$
\Comment{policy update}

        \State $\phi_{k+1}\gets\alpha\phi_k+(1-\alpha)\theta_{k+1}$
        \Comment{EMA teacher update}
    \EndFor
    \State \Return $\pi_{\theta_K}$
  \end{algorithmic}
\end{algorithm}
\end{minipage}
\end{wrapfigure}

The resulting $\Omega_x^{(k)}$ aggregates procedural knowledge discovered across successful trajectories while retaining the constituent Atoms as distinct units. It is then shared across all trajectories of task $x$ to provide reusable guidance over different interaction histories, including both successful and unsuccessful ones.

\subsection{Atom-Guided Self-Distillation}
\label{sec:atom-self-distillation}

The second process completes each self-improving loop by distilling the procedural knowledge $\Omega_x^{(k)}$ extracted from the current student back into its policy.
Conditioned on this shared Atom set, the teacher provides token-level procedural supervision over the student's on-policy interaction histories.
Algorithm~\ref{alg:atomsd} summarizes the complete self-improving loop executed at each training iteration.

\paragraph{Atom-Conditioned Guidance}

Let $\phi_k$ denote the teacher parameters at iteration $k$. For each trajectory $\tau_i\in\mathcal{T}_x^{(k)}$ and the $j$-th student-generated token $y_{ij}$ in $\tau_i$, let $h_{ij}$ denote its preceding token-level interaction history. Given the shared task-level Atom set $\Omega_x^{(k)}$ and the teacher prompting template shown in Figure~\ref{fig:prompt_teacher}, the teacher's Atom-conditioned token-level support for $y_{ij}$ is defined as
\begin{equation}
    q_{ij}^{\mathrm{Atom},(k)}
    =
    \pi_{\phi_k}
    \left(
        y_{ij}\mid h_{ij},\Omega_x^{(k)}
    \right).
    \label{eq:atom-teacher-score}
\end{equation} 
For each Atom $\omega\in\Omega_x^{(k)}$, the intent $\iota_\omega$ and applicability conditions $\mathcal{C}_\omega$ provide the basis for identifying relevant Atoms from the current history $h_{ij}$, while $\mathcal{P}_\omega$ and $\mathcal{B}_\omega$ provide the corresponding procedural guidance and boundaries.
Conditioning jointly on $h_{ij}$ and $\Omega_x^{(k)}$ allows the teacher to draw on relevant Atom guidance while providing history-specific token-level supervision, enabling compositional procedural guidance across multiple Atoms.

\paragraph{Joint Optimization}
We convert the Atom-conditioned support from a teacher derived from the student into a token-level internalization signal:
\begin{equation}
\fontsize{10}{9}\selectfont
\begin{aligned}
    A_{ij}^{\mathrm{Atom},(k)}
    &=
    \operatorname{sg}\!\left[
        \log q_{ij}^{\mathrm{Atom},(k)}
        -
        \log \pi_\theta(y_{ij}\mid h_{ij})
    \right],
\end{aligned}
\label{eq:atom-advantage}
\end{equation}
where $\operatorname{sg}(\cdot)$ denotes stop-gradient. This signal captures the difference between the teacher's support under the extracted procedural knowledge and the student's corresponding unguided probability.

In parallel, environment feedback provides a trajectory-level learning signal. For the $G$ trajectories in $\mathcal{T}_x^{(k)}$, we compute the GRPO advantage~\cite{grpo} as $A_i^{\mathrm{RL},(k)}=(R(\tau_i)-\overline{R}_x^{(k)})/\sigma_x^{(k)}$, where $\overline{R}_x^{(k)}$ and $\sigma_x^{(k)}$ denote the within-task reward mean and standard deviation. The policy ratio is $r_{ij}^{(k)}(\theta)=\pi_\theta(y_{ij}\mid h_{ij})/\pi_{\theta_k}(y_{ij}\mid h_{ij})$.
Both learning signals use the same PPO-style clipped surrogate loss~\cite{schulman2017proximalpolicyoptimizationalgorithms}:
\begin{equation}
\fontsize{10}{9}\selectfont
\begin{aligned}
    \ell_{\mathrm{clip}}\!\left(r_{ij}^{(k)}(\theta),A\right)
    &=
    -\min\Bigl\{
        r_{ij}^{(k)}(\theta)A,
        \operatorname{clip}\!\left(
        r_{ij}^{(k)}(\theta),1-\epsilon,1+\epsilon
        \right)A
    \Bigr\},
\end{aligned}
\label{eq:clipped-surrogate}
\end{equation}
where $\operatorname{clip}(\cdot)$ constrains the policy ratio to $[1-\epsilon,1+\epsilon]$, and $A$ denotes the corresponding reward or Atom-guided distillation signal.
Combining the two signals, the task-level objective is
\begin{equation}
\fontsize{10}{9}\selectfont
\begin{aligned}
\mathcal{L}_x^{(k)}(\theta)
&=
\frac{1}{\sum_{i=1}^{G}L_i}
\sum_{i=1}^{G}\sum_{j=1}^{L_i}
\Big[
\ell_{\mathrm{clip}}\!\left(
r_{ij}^{(k)}(\theta),A_i^{\mathrm{RL},(k)}
\right) 
+\lambda\,
\ell_{\mathrm{clip}}\!\left(
r_{ij}^{(k)}(\theta),
A_{ij}^{\mathrm{Atom},(k)}
\right)
\Big],
\end{aligned}
\label{eq:joint-loss}
\end{equation}
where $L_i$ counts valid student-generated tokens in $\tau_i$, and $\lambda$ controls the strength of distillation. The batch objective aggregates these task-level losses according to their valid token counts. Optimizing this objective updates the student from $\pi_{\theta_k}$ to $\pi_{\theta_{k+1}}$, progressively internalizing the procedural knowledge represented by $\Omega_x^{(k)}$ into the policy.

\paragraph{Teacher Update}

The teacher and student are initialized identically, $\phi_0=\theta_0$. After optimizing each training iteration, the teacher is updated by exponential moving average (EMA):
\begin{equation}
    \phi_{k+1}
    =
    \alpha\phi_k
    +
    (1-\alpha)\theta_{k+1},
    \label{eq:teacher-ema}
\end{equation}
where $\alpha$ denotes the EMA decay.
The updated student $\pi_{\theta_{k+1}}$ then serves as the basis for procedural abstraction in the next iteration. Repeating this loop continually externalizes procedural knowledge from the evolving learner and distills it back into the policy, allowing procedural knowledge and the policy to co-evolve throughout training.

\begin{table*}[t]
\centering
\small
\setlength{\tabcolsep}{5pt}
\renewcommand{\arraystretch}{1.05}
\begin{adjustbox}{max width=\linewidth}
\begin{tabular}{lccccccc}
\toprule
\multirow{2}{*}[-0.5\dimexpr\aboverulesep+\belowrulesep+\cmidrulewidth\relax]{\textbf{Model / Method}} & \multicolumn{2}{c}{\textbf{ALFWorld}} & \multicolumn{2}{c}{\textbf{ScienceWorld}} & \multirow{2}{*}[-0.5\dimexpr\aboverulesep+\belowrulesep+\cmidrulewidth\relax]{\textbf{WebShop}} & \multirow{2}{*}[-0.5\dimexpr\aboverulesep+\belowrulesep+\cmidrulewidth\relax]{\textbf{StableToolBench}} & \multirow{2}{*}[-0.5\dimexpr\aboverulesep+\belowrulesep+\cmidrulewidth\relax]{\textbf{Avg.}} \\
\cmidrule(lr){2-3}\cmidrule(lr){4-5}
& \textit{Seen} & \textit{Unseen} & \textit{Seen} & \textit{Unseen} & & & \\
\midrule
DeepSeek-V4-Flash\,\deepseekicon & 78.57 & 73.13 & 73.25 & 68.22 & 60.25 & 48.52 & 66.99 \\
DeepSeek-V4-Pro\,\deepseekicon & 85.71 & 85.82 & 76.80 & 76.43 & 60.95 & 52.66 & 73.06 \\
GLM-5.3\,\glmicon & 93.57 & 91.79 & 80.45 & 72.69 & 70.88 & 64.50 & 78.98 \\
GPT-5.5\,\chatgpticon & 97.14 & 100.00 & 87.16 & 86.22 & 71.00 & 58.58 & 83.35 \\
\midrule
\multicolumn{8}{l}{\cellcolor{gray!10}\textit{Qwen3.5-4B}\,\qwenicon} \\
\freezemark\,ReAct & 50.71 & 50.00 & 27.49 & 19.97 & 21.54 & 50.30 & 36.67 \\
\freezemark\,Reflexion & 69.29 & 63.43 & 36.79 & 29.33 & 28.04 & 68.64 & 49.25 \\
\trainmark\,OPSD & \thirdscore{80.00} & \secondscore{79.85} & \thirdscore{40.74} & 30.92 & 50.40 & 63.31 & 57.54 \\
\trainmark\,GRPO & 79.29 & \thirdscore{79.10} & 24.81 & 23.08 & \secondscore{63.65} & 72.19 & 57.02 \\
\trainmark\,GRPO+OPSD & \secondscore{85.71} & 78.36 & 32.46 & \thirdscore{32.42} & \bestscore{64.14} & \secondscore{84.02} & \secondscore{62.85} \\
\trainmark\,SDPO & 46.43 & 41.04 & 5.40 & 4.69 & 0.67 & 2.37 & 16.77 \\
\trainmark\,Skill-SD & 78.57 & 72.39 & 26.97 & 24.36 & 56.34 & 71.01 & 54.94 \\
\trainmark\,SDAR & 76.43 & 74.63 & \secondscore{46.74} & \secondscore{42.28} & 47.02 & \thirdscore{76.33} & \thirdscore{60.57} \\
\trainmark\,\textbf{AtomSD (Ours)} & \bestscore{93.57} & \bestscore{84.33} & \bestscore{55.73} & \bestscore{54.13} & \thirdscore{60.58} & \bestscore{84.62} & \bestscore{72.16}\,\gain{9.31} \\
\midrule
\multicolumn{8}{l}{\cellcolor{gray!10}\textit{Qwen3.5-9B}\,\qwenicon} \\
\freezemark\,ReAct & 56.43 & 51.49 & 29.22 & 25.56 & 39.75 & 52.07 & 42.42 \\
\freezemark\,Reflexion & 70.00 & 71.64 & 48.48 & 43.31 & 38.84 & 70.41 & 57.11 \\
\trainmark\,OPSD & \thirdscore{87.14} & \thirdscore{85.82} & 60.70 & 57.96 & 58.53 & 68.05 & 69.70 \\
\trainmark\,GRPO & 82.14 & 82.09 & \thirdscore{69.09} & \secondscore{68.33} & 50.56 & 71.01 & 70.54 \\
\trainmark\,GRPO+OPSD & 77.86 & 78.36 & \secondscore{69.15} & \thirdscore{62.08} & \bestscore{62.58} & \thirdscore{85.21} & \secondscore{72.54} \\
\trainmark\,SDPO & 65.71 & 60.45 & 16.12 & 13.30 & 0.97 & 11.83 & 28.06 \\
\trainmark\,Skill-SD & 80.00 & 83.58 & 61.82 & 60.71 & \thirdscore{59.57} & \secondscore{88.76} & \thirdscore{72.41} \\
\trainmark\,SDAR & \secondscore{87.86} & \bestscore{91.04} & 30.45 & 28.81 & 58.73 & 78.70 & 62.60 \\
\trainmark\,\textbf{AtomSD (Ours)} & \bestscore{88.57} & \secondscore{86.57} & \bestscore{70.68} & \bestscore{68.75} & \secondscore{60.89} & \bestscore{92.31} & \bestscore{77.96}\,\gain{5.42} \\
\bottomrule
\end{tabular}
\end{adjustbox}
\caption{Main results across four benchmarks. \freezemark{} denotes methods requiring no additional training; \trainmark{} denotes methods that update model parameters. Higher is better in every column. Avg. is the average of the six reported scores; green arrows (\textcolor{GainGreen}{$\boldsymbol{\uparrow}$}) indicate our absolute gains over the highest baseline Avg. within each backbone. \protect\colorbox{ResultBest}{\textbf{Best}}, \protect\colorbox{blue!5}{\underline{Second-best}}, and \protect\colorbox{ResultThird}{Third-best} scores are highlighted within each backbone.}
\label{tab:main-results}
\vspace{-1em}
\par\medskip

\begin{minipage}[t]{1\textwidth}
    \vspace{1pt}
    \centering
    \includegraphics[width=\linewidth]{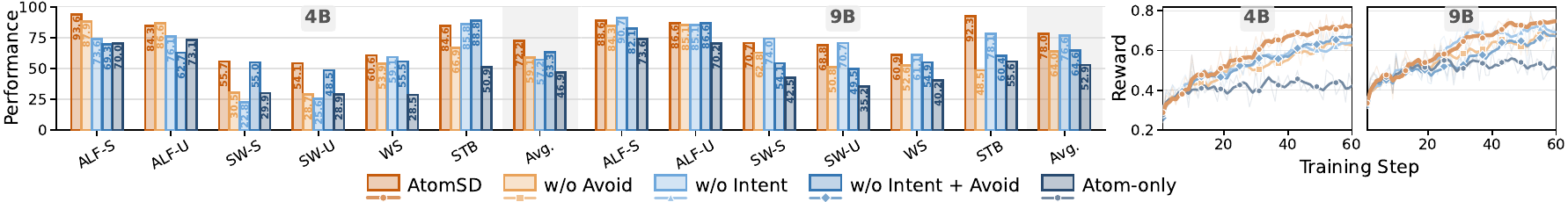}
    \captionof{figure}{Ablation of Atom components on Qwen3.5-4B and Qwen3.5-9B. \textbf{Left}: Final benchmark performance. \textbf{Right}: Training rewards after removing Intent, Avoid, or both, alongside the Atom-only setting.}
    \label{fig:ablation_summary_components}
\end{minipage}
\vspace{-1em}
\end{table*}

\section{Experiments}

\subsection{Experiment Settings}

\paragraph{Benchmarks}
We evaluate agents' ability to follow and execute multi-step procedures across four benchmarks. \textbf{ALFWorld (ALF)}~\cite{shridhar2021alfworldaligningtextembodied} assesses household planning and execution in text-based environments. \textbf{ScienceWorld (SW)}~\cite{wang2022scienceworldagentsmarter5th} requires scientific reasoning through multi-step experimentation. \textbf{WebShop (WS)}~\cite{yao2023webshopscalablerealworldweb} evaluates product search and purchasing under user constraints. \textbf{StableToolBench (STB)}~\cite{guo2025stabletoolbenchstablelargescalebenchmarking} assesses tool use through a virtual API environment. We report seen and unseen results separately for ALFWorld and ScienceWorld, denoted by the suffixes \textbf{-S} and \textbf{-U}, alongside test results for WebShop and StableToolBench. Detailed descriptions of the datasets and evaluation subsets are provided in Appendix~\ref{sec:dataset-details}.

\paragraph{Models}
We use Qwen3.5-4B and Qwen3.5-9B~\cite{qwen2026qwen35} as backbone models. We additionally evaluate four strong API-based models:
DeepSeek-V4-Flash and DeepSeek-V4-Pro~\cite{deepseek2026v4pro}, GLM-5.3~\cite{glm5team2026glm5}, and GPT-5.5~\cite{openai2026gpt55}. 

\paragraph{Baselines}
All models interact with the environments using ReAct~\cite{yao2023react}. API-based models are evaluated with ReAct alone.
We compare AtomSD with the following categories of methods: 
(1) \textbf{Prompting}: the initial model with ReAct alone and Reflexion~\cite{Reflexion}, which additionally incorporates textual reflections from previous attempts without parameter updates;
(2) \textbf{Reward-based RL}: GRPO~\mbox{\cite{grpo}},
which learns from task rewards;
(3) \textbf{Trajectory-guided self-distillation}:
OPSD~\cite{opsd}, SDPO~\cite{sdpo}, and GRPO+OPSD,
which use successful demonstrations as privileged teacher context
in our setting, with GRPO+OPSD additionally incorporating task rewards;
and (4) \textbf{Skill-guided distillation}:
Skill-SD~\cite{skill_sd} and SDAR~\cite{sdar},
which combine reward optimization with teacher guidance from skills constructed by strong external LLMs.
Together, these baselines provide a comprehensive comparison across inference-time reflection, reward-based learning, and trajectory- or skill-guided distillation.
We set $\lambda=0.001$, $\alpha=0.95$, and $\epsilon=0.2$ for AtomSD. Detailed training and implementation settings are provided in Appendix~\ref{sec:implementation-details}.

\paragraph{Evaluation Metrics}
For training-based methods, we evaluate checkpoints at the same training step. We report success rate on ALFWorld, average task score on ScienceWorld and WebShop, and solved rate on StableToolBench using the official Final Answer Completeness (FAC)~\cite{guo-etal-2025-stabletoolbench-mirror-api}. All scores use a 0--100 scale.

\subsection{Main Results}
Table~\ref{tab:main-results} summarizes the main results, with four key findings.
\noindent\textbf{(1) AtomSD improves beyond reward- and demonstration-guided training.}
AtomSD consistently outperforms GRPO, OPSD, and GRPO+OPSD in average performance at both model scales, exceeding the strongest baseline by 9.31 and 5.42 points at 4B and 9B, respectively. This highlights the value of Procedural Atoms beyond reward-only learning, demonstration-guided distillation, and their combination.
\noindent\textbf{(2) AtomSD provides broader gains than skill-level distillation.}
AtomSD outperforms Skill-SD and SDAR on most tasks at both scales, with particularly pronounced gains on ScienceWorld. These results suggest that even skill libraries constructed by stronger external models do not necessarily provide more effective guidance, highlighting the value of fine-grained Procedural Atoms.
\noindent\textbf{(3) Improvements generalize to unseen tasks.}
AtomSD leads on ScienceWorld unseen at both model scales, with improvements also extending to ALFWorld unseen. These results suggest that procedural knowledge learned through AtomSD can generalize beyond seen tasks.
\noindent\textbf{(4) AtomSD is competitive with strong LLMs without inference-time reflection.}
AtomSD surpasses Reflexion by 22.91 and 20.85 points at 4B and 9B, respectively. AtomSD-9B further exceeds DeepSeek-V4-Pro and trails GPT-5.5 by only 5.39 points in average score, suggesting that internalizing Procedural Atom guidance can reduce reliance on inference-time reflection while narrowing the gap to substantially stronger LLMs. We further provide qualitative case studies in Appendix~\ref{sec:procedural-dependency-patterns}.

\subsection{Effects of Atom Components and Joint Optimization}

Figure~\ref{fig:ablation_summary_components} examines the contributions of Atom components and joint reward optimization.
\noindent\textbf{(1) The complete Atom representation provides more effective guidance.}
Including Intent improves the average score by 15.0 points at 4B and 1.4 points at 9B, while the complete representation outperforms removing both Intent and Avoid by 8.9 and 13.4 points, respectively, with similar trends observed in the training rewards.
These results highlight the value of organizing procedural guidance within the complete Atom representation.
\noindent\textbf{(2) Explicit avoidance boundaries improve performance at both scales.}
Removing Avoid reduces the average score by 13.1 points at 4B and 14.0 points at 9B. This further highlights the importance of explicitly representing what to avoid, rather than distilling successful operations alone.
\noindent\textbf{(3) Joint optimization strengthens Atom-guided learning.}
Combining Atom-guided distillation with task rewards improves average performance by 25.3 and 25.1 points over Atom-only training at 4B and 9B, respectively. 
This supports the design choice of jointly internalizing procedural guidance while optimizing for task completion. Overall, these results show that effective Atom-guided learning depends on both a complete procedural representation and joint optimization with task rewards.

\begin{figure*}[t]
    \centering
    \begin{minipage}[t]{0.29\textwidth}
        \vspace{-125pt}
        \centering
        \includegraphics[width=\linewidth]{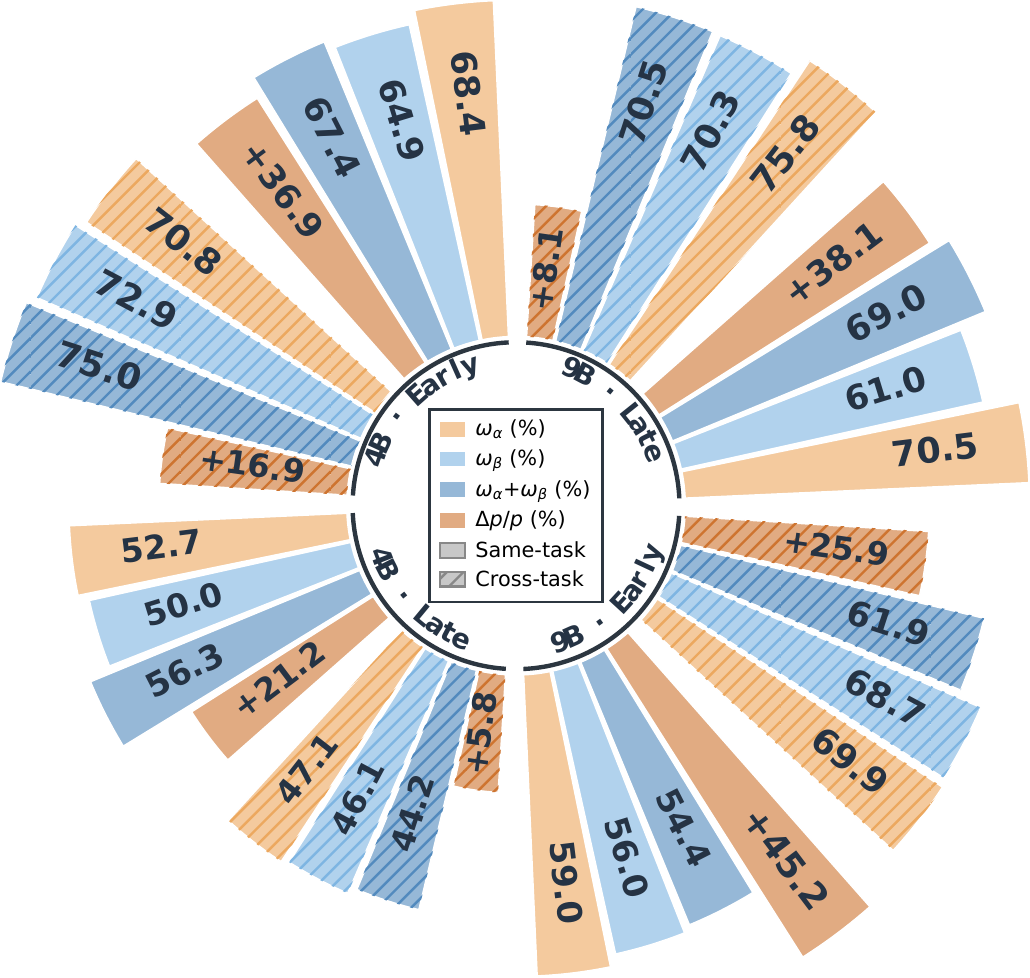}
        \caption{
        Procedural Atom internalization analysis on Qwen3.5-4B/9B across \textit{Early}/\textit{Late} training and same-/cross-task histories, reporting action-support change and guidance-direction agreement.
        }
        \label{fig:state-transfer}
        \par\smallskip
    \end{minipage}
    \hfill
    \begin{minipage}[t]{0.7\textwidth}
        \centering
        \includegraphics[width=\linewidth]{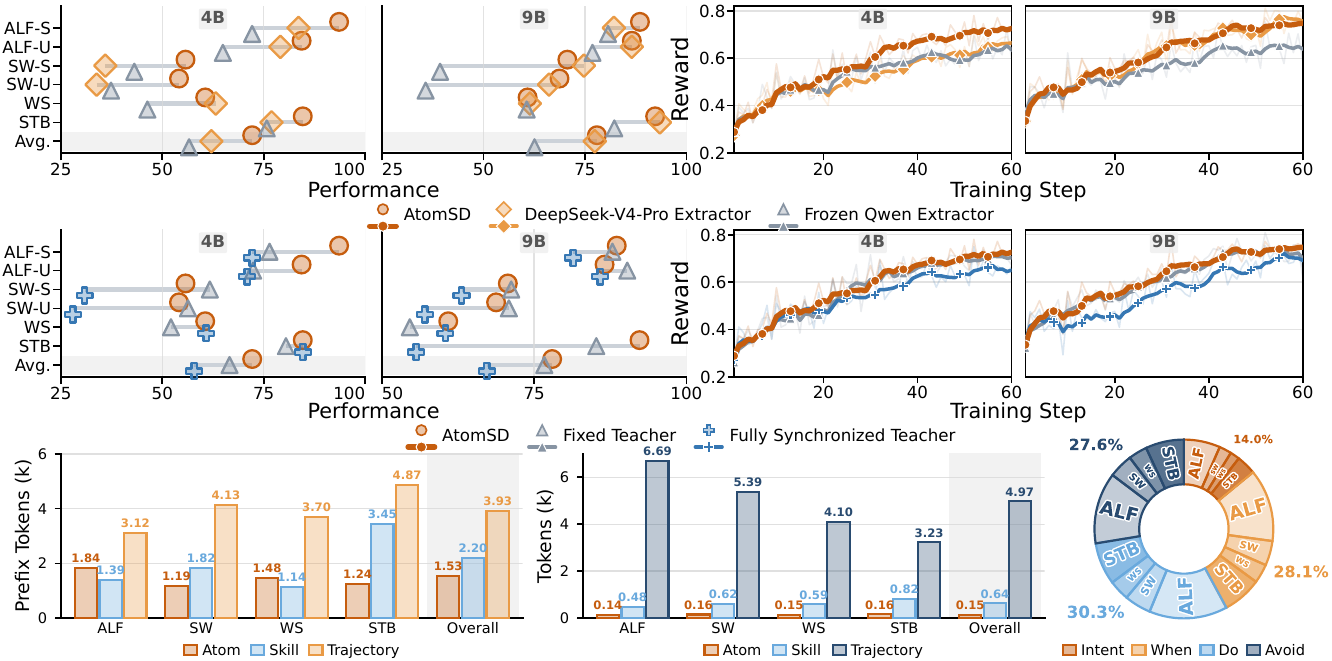}
        \par\smallskip
        \caption{
Analysis of Procedural Atom evolution, teacher updates, and representation.
\textbf{Top}: Extractor comparison.
\textbf{Middle}: Teacher-update comparison.
\textbf{Bottom left}: Length comparison of teacher prefixes (left) and individual Atoms, skills, and student trajectories (right); \textit{Overall} averages all samples across the four datasets.
\textbf{Bottom right}: Atom component count distribution.
}
    \label{fig:atom-analysis}
    \end{minipage}
\end{figure*}

\subsection{Internalization and Co-Evolution}

\paragraph{Internalization and Composition}
To examine whether the student internalizes Atom guidance, we consider two relevant Procedural Atoms, $\omega_\alpha$ and $\omega_\beta$, and evaluate guidance from each individually and in combination. We use GPT-5.5 to identify supported actions and their preceding interaction histories. This analysis covers both same-task and cross-task histories at two training stages, \textit{Early} and \textit{Late}. Within each stage, we use a fixed training interval and compare the student checkpoints at its beginning and end using two metrics: (1) \emph{maximum-token support change}, measuring the relative probability increase of the most-improved action token without Atoms, and (2) \emph{guidance-direction agreement}, measuring whether Atom-favored tokens change in the same direction after training without Atoms. Detailed definitions and results are provided in Appendix~\ref{sec:atom-internalization}.
As shown in Figure~\ref{fig:state-transfer}, during \textit{Early} training, maximum-token support ($\Delta p/p$) increases by 36.9\% for 4B and 45.2\% for 9B on same-task histories. Agreement with composed guidance from $\omega_\alpha$ and $\omega_\beta$ reaches 75.0\% on cross-task histories for \textit{Early} 4B and 70.5\% for \textit{Late} 9B. These results support effective internalization of both individual and composed Atom guidance, including across tasks; see Appendix~\ref{sec:atom-composition-case} for a concrete composition case.

\paragraph{Co-Evolution}
To assess whether procedural knowledge benefits from co-evolving with the learner, we replace the evolving student extractor with either DeepSeek-V4-Pro or the frozen initial Qwen backbone. Figure~\ref{fig:atom-analysis} (top) shows that a stronger extractor alone does not yield better performance: DeepSeek-V4-Pro trails AtomSD at 4B and is comparable at 9B. Meanwhile, the evolving student consistently outperforms the frozen extractor in evaluation performance and late-stage rewards at both scales. These results support AtomSD's co-evolving design: learner improvement refreshes procedural knowledge for distillation, which in turn supports further policy improvement.

\paragraph{Atom Representation}
To characterize the granularity and compactness of Procedural Atoms, Figure~\ref{fig:atom-analysis} (bottom left) compares lengths using trajectories matched by task and training step across methods. 
Overall, Atom guidance requires 1.53k prefix tokens, compared with 2.20k for skills and 3.93k for trajectories. At the unit level, individual Atoms average only 0.15k tokens, substantially smaller than skills (0.64k) and trajectories (4.97k). 
Figure~\ref{fig:atom-analysis} (bottom right) further shows a relatively balanced distribution of the four Atom components across datasets throughout training.
These results show that Procedural Atoms provide compact, fine-grained guidance across multiple procedural aspects, supporting selective reuse and composition.

\begin{figure*}[t]
    \centering
    \begin{minipage}[t]{0.57\textwidth}
        \vspace{0pt}
        \centering
        \includegraphics[
            width=\linewidth,
            height=4.2cm,
            keepaspectratio
        ]{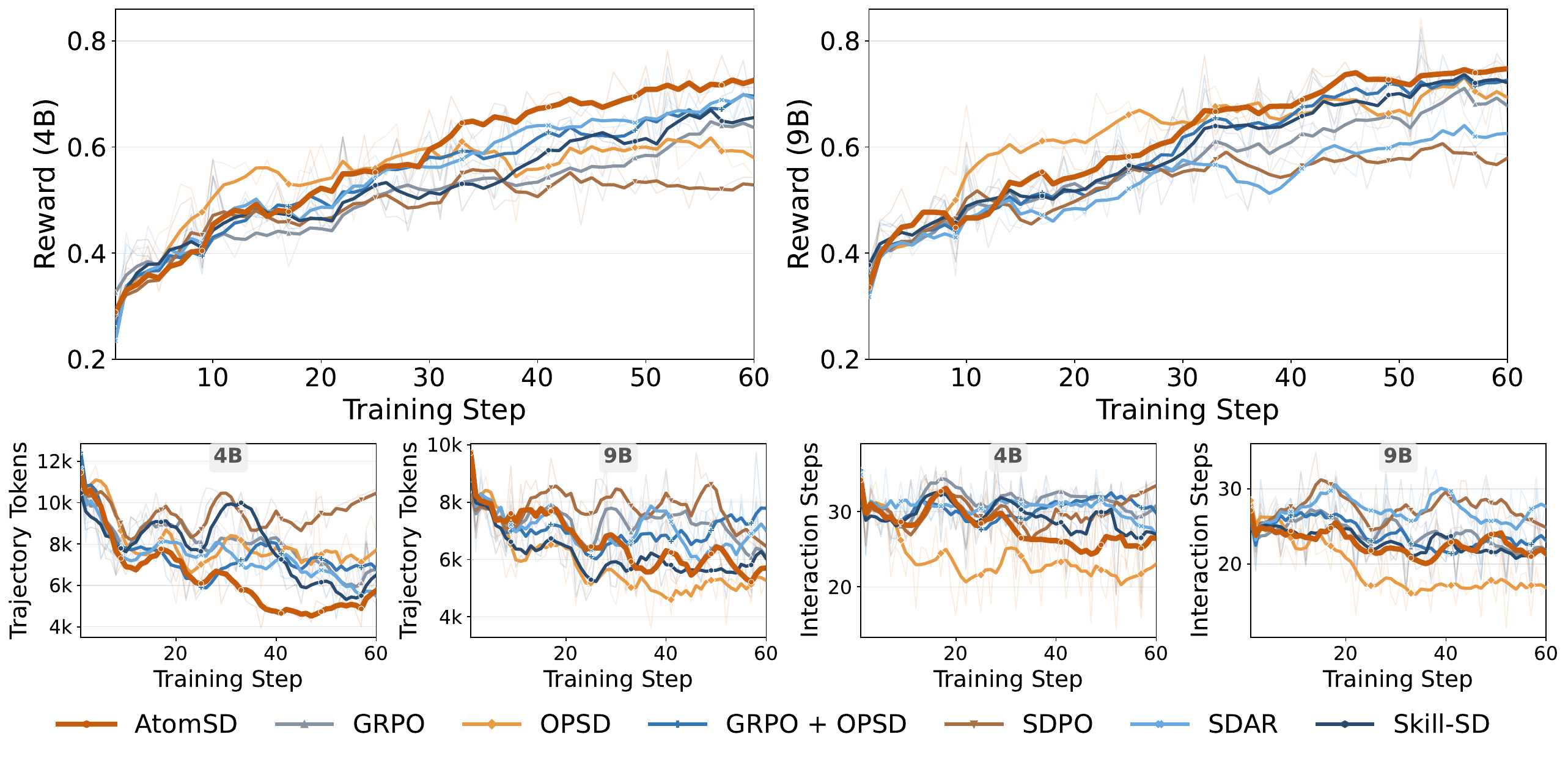}
        \caption{Training dynamics of AtomSD and baselines on Qwen3.5-4B and Qwen3.5-9B. \textbf{Top}: Training rewards. \textbf{Bottom}: Trajectory token counts (left) and interaction steps (right).}
        \label{fig:reward-trajectory-curves}
    \end{minipage}%
    \hfill%
    \begin{minipage}[t]{0.415\textwidth}
        \vspace*{1.1mm}
        \centering
        \includegraphics[
            width=\linewidth,
            height=4.2cm,
            keepaspectratio
        ]{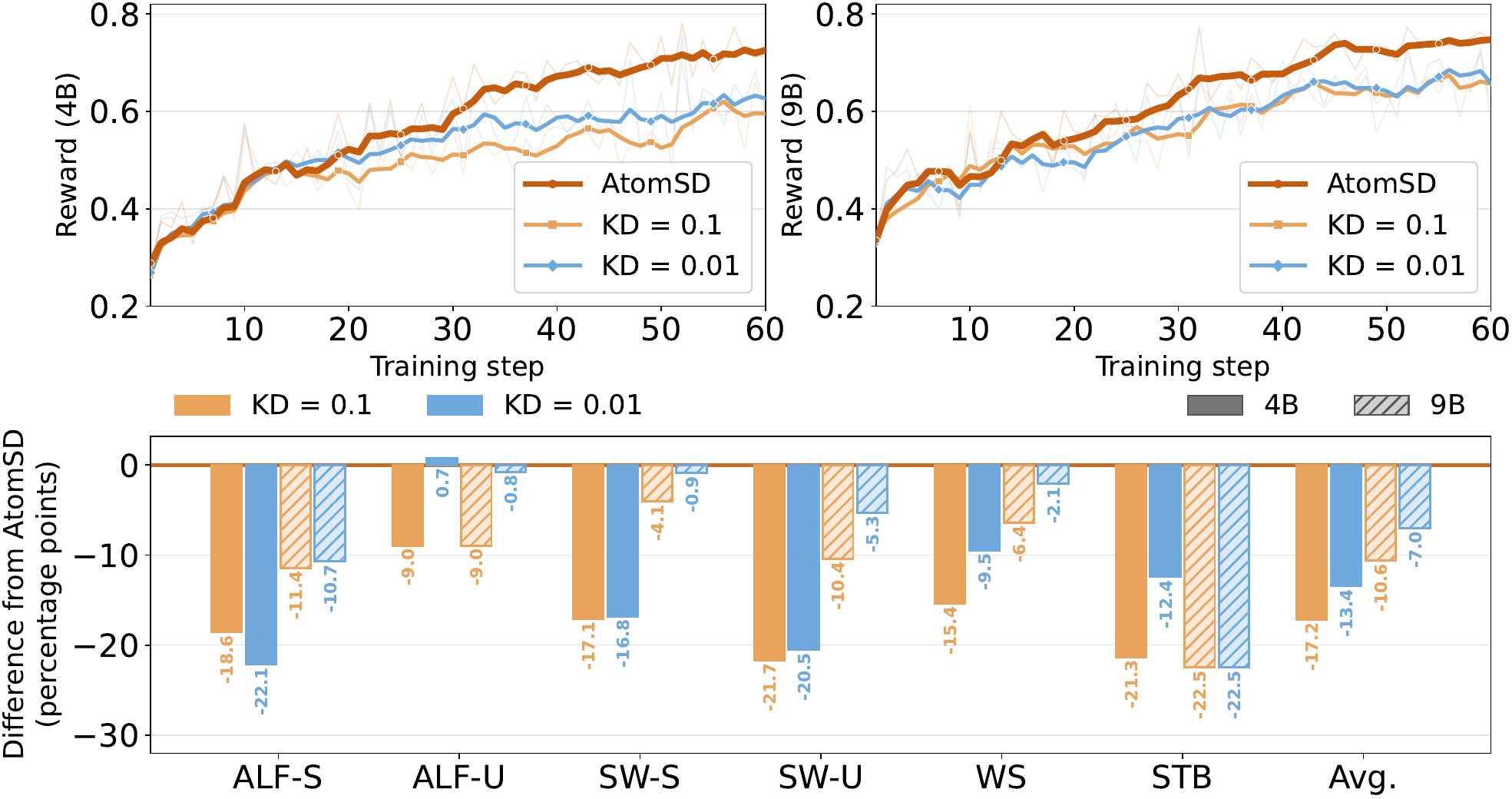}
        \caption{Distillation coefficient analysis on Qwen3.5-4B and Qwen3.5-9B. \textbf{Top}: Training rewards. \textbf{Bottom}: Final evaluation score differences from AtomSD.}
        \label{fig:ablation_summary_kd}
    \end{minipage}
\end{figure*}

\subsection{Analyses}

\paragraph{Training Dynamics Analysis}
Figure~\ref{fig:reward-trajectory-curves} (top) shows that AtomSD's advantage emerges as training progresses. OPSD improves rapidly early on at both model scales, but its gains subsequently plateau, whereas AtomSD continues to improve and reaches the highest reward toward the end. Figure~\ref{fig:reward-trajectory-curves} (bottom) further shows that these gains accompany relatively low token usage and fewer interaction steps during later training at both scales. These results show that Procedural Atoms improve performance with more compact trajectories.

\paragraph{Distillation Strength Analysis}
Figure~\ref{fig:ablation_summary_kd} shows that increasing $\lambda$ from 0.001 to 0.01 or 0.1 consistently reduces late-stage training rewards and average evaluation performance at both model scales. This suggests that Atom-guided distillation is more effective when properly balanced with reward optimization rather than simply strengthened, consistent with recent findings on self-distillation in agentic RL~\cite{sdar,kaur2026rethinkingonpolicyselfdistillationthinking}.

\paragraph{Teacher Update Analysis}
Figure~\ref{fig:atom-analysis} (middle) compares EMA with fixed and fully synchronized teachers. EMA achieves the highest average evaluation performance at both model scales; the fixed teacher remains competitive, while full synchronization yields lower late-stage training rewards at both scales. These results suggest that EMA balances preserving prior teacher knowledge with incorporating student progress.

\section{Related Work}
\paragraph{Experiential Memory and Procedural Knowledge}
Experience-based agents transform interaction histories into reflections, insights, and reasoning memories for reuse in subsequent decisions~\cite{Reflexion,zhao2024expelllmagentsexperiential,ouyang2026reasoningbankscalingagentselfevolving}, while others leverage embeddings, KV cache, or special tokens to represent memories~\cite{yu2025srkiscalablerealtimeknowledge,yu2026premlearningpreserverefresh,wu2026tokmemonetokenproceduralmemory}.
Furthermore, prior work organizes histories into reusable workflows, conditional procedures, and explicit rules, enabling more selective reuse across interactions~\cite{Agent_Workflow_Memory,AutoGuide,lin2026imitateboundaryawareskillmemory,yu2026factualknowledgebenchmarkinglearning,mi2026skillpro}. A complementary line of work models environment dynamics and action consequences to support planning~\cite{qiao2024wkm,kim2025coex,liu2026comap}. AtomSD instead uses procedural abstractions as privileged context. Condition-aware Atoms guide policy learning and are progressively internalized into model parameters, without requiring external guidance at inference.

\paragraph{On-Policy Distillation}
On-policy distillation applies teacher supervision to student-generated trajectories, internalizing privileged knowledge into model parameters~\cite{agarwal2024onpolicy,opsd,sdpo,ye2026onpolicycontextdistillationlanguage,wu2026contrastive}. Privileged knowledge can be supplied as explicit textual context or encoded into learned latent representations~\cite{liu2026proprietary,ye2026onpolicycontextdistillationlanguage,zhang2026latentonpolicyselfdistillation}. AtomSD uses independently applicable Procedural Atoms as the unit of privileged supervision, enabling guidance to be composed according to the student's current interaction history rather than being tied to a specific demonstration trajectory or workflow.

\paragraph{Skill-Guided Agent Learning}
Skill-augmented reinforcement learning optimizes agents' ability to exploit and improve external skill libraries~\cite{wang-etal-2026-reinforcement,xia2026skillrlevolvingagentsrecursive,chen2026decoevo}. Skill-conditioned distillation instead uses skill guidance as teacher-side context to train policies that operate without external skill inputs at inference~\cite{sdar,skill_sd,yang2026opidonpolicyskilldistillation,lv2026pcsd}. AtomSD constructs procedural guidance online from the evolving student's successful trajectories, without relying on a preconstructed skill library or a separate external model for knowledge extraction. By continually decomposing trajectories into independently applicable Atoms rather than bundled skills, it supports selective reuse and flexible composition while allowing procedural knowledge to co-evolve with the learner.

\section{Conclusion}

This paper presents AtomSD, a self-improving framework that transforms successful on-policy trajectories into Procedural Atoms and distills their guidance back into the policy, allowing procedural knowledge to co-evolve with the learner. Comprehensive experiments and analyses show that AtomSD consistently improves agent performance, while fine-grained Procedural Atoms support selective reuse and flexible composition across interaction histories. Our results highlight atomized procedural knowledge as an effective bridge between raw interaction histories and parametric policy learning.

\bibliographystyle{colm2026_conference}
\bibliography{custom}

@String{Computing = "Computing" }

@String{Computer = "{IEEE} Computer" }

@String{Chelsea = "Chelsea" }

@ArtifactSoftware{R,
    title = {R: A Language and Environment for Statistical Computing},
    author = {{R Core Team}},
    organization = {R Foundation for Statistical Computing},
    address = {Vienna, Austria},
    year = {2019},
    url = {https://www.R-project.org/},
}

@misc{deepseek2026v4pro,
      title={DeepSeek-V4: Towards Highly Efficient Million-Token Context Intelligence}, 
      author={DeepSeek-AI and Anyi Xu and Bangcai Lin and Bing Xue and Bingxuan Wang and Bingzheng Xu and Bochao Wu and Bowei Zhang and Chaofan Lin and Chen Dong and Chenchen Ling and Chengda Lu and Chenggang Zhao and Chengqi Deng and Chengyu Hou and Chenhao Xu and Chenze Shao and Chong Ruan and Conner Sun and Damai Dai and Daya Guo and Dejian Yang and Deli Chen and Donghao Li and Dongjie Ji and Erhang Li and Fang Wei and Fangyun Lin and Fangzhou Yuan and Feiyu Xia and Fucong Dai and Guangbo Hao and Guanting Chen and Guoai Cao and Guolai Meng and Guowei Li and Han Yu and Han Zhang and Hanwei Xu and Hao Li and Haofen Liang and Haoling Zhang and Haoming Luo and Haoran Wei and Haotian Yuan and Haowei Zhang and Haowen Luo and Haoyu Chen and Haozhe Ji and Hengqing Zhang and Honghui Ding and Hongxuan Tang and Huanqi Cao and Huazuo Gao and Hui Qu and Hui Zeng and J Yang and JQ Zhu and Jia Luo and Jia Song and Jia Yu and Jialiang Huang and Jialu Cai and Jian Liang and Jiangting Zhou and Jiasheng Ye and Jiashi Li and Jiaxin Xu and Jiewen Hu and Jieyu Yang and Jin Chen and Jin Yan and Jingchang Chen and Jingli Zhou and Jingting Xiang and Jingyang Yuan and Jingyuan Cheng and Jingzi Zhou and Jinhua Zhu and Jiping Yu and Joseph Sun and Jun Ran and Junguang Jiang and Junjie Qiu and Junlong Li and Junmin Zheng and Junxiao Song and Kai Dong and Kaige Gao and Kang Guan and Kexing Zhou and Kezhao Huang and Kuai Yu and Lean Wang and Lecong Zhang and Lei Wang and Leyi Xia and Li Zhang and Liang Zhao and Lihua Guo and Lingxiao Luo and Linwang Ma and Linyan Zhu and Litong Wang and Liyu Cai and Liyue Zhang and Longhao Chen and MS Di and MY Xu and Max Mei and Miaojun Wang and Mingchuan Zhang and Minghua Zhang and Minghui Tang and Mingming Li and Mingxu Zhou and Minmin Han and Ning Wang and Panpan Huang and Panpan Wang and Peixin Cong and Peiyi Wang and Peng Zhang and Qiancheng Wang and Qihao Zhu and Qingyang Li and Qinyu Chen and Qiushi Du and Qiwei Jiang and Rui Tian and Ruifan Xu and Ruijie Lu and Ruiling Xu and Ruiqi Ge and Ruisong Zhang and Ruizhe Pan and Runji Wang and Runqian Chen and Runqiu Yin and Runxin Xu and Ruomeng Shen and Ruoyu Zhang and Ruyi Chen and SH Liu and Shanghao Lu and Shangmian Sun and Shangyan Zhou and Shanhuang Chen and Shaofei Cai and Shaoheng Nie and Shaoqing Wu and Shaoyuan Chen and Shengding Hu and Shengyu Liu and Shiqiang Hu and Shirong Ma and Shiyu Wang and Shuiping Yu and Shunfeng Zhou and Shuting Pan and Shuying Yu and Songyang Zhou and Tao Ni and Tao Yun and Tian Jin and Tian Pei and Tian Ye and Tianle Lin and Tianran Ji and Tianyi Cui and Tianyuan Yue and Tingting Yu and Tun Wang and W Zhang and WL Xiao and Wangding Zeng and Wei An and Weilin Zhao and Wen Liu and Wenfeng Liang and Wenjie Pang and Wenjing Luo and Wenjing Yao and Wenjun Gao and Wenkai Yang and Wenlve Huang and Wenqing Hou and Wentao Zhang and Wenting Ma and Xi Gao and Xiang He and Xiangwen Wang and Xianzu Wang and Xiao Bi and Xiaodong Liu and Xiaohan Wang and Xiaokang Chen and Xiaokang Zhang and Xiaotao Nie and Xiaowen Sun and Xiaoxiang Wang and Xin Cheng and Xin Liu and Xin Xie and Xingchao Liu and Xingchen Liu and Xingkai Yu and Xingyou Li and Xinyu Yang and Xinyu Zhang and Xu Chen and Xuanyu Wang and Xuecheng Su and Xueyin Chen and Xuheng Lin and Xuwei Fu and YC Yan and YQ Wang and YW Ma and Yanfeng Luo and Yang Zhang and Yanhong Xu and Yanru Ma and Yanwen Huang and Yao Li and Yao Li and Yao Xu and Yao Zhao and Yaofeng Sun and Yaohui Wang and Yi Qian and Yi Shao and Yi Yu and Yichao Zhang and Yifan Ding and Yifan Shi and Yijia Wu and Yiliang Xiong and Yiling Ma and Ying He and Ying Tang and Ying Zhou and Yingjia Luo and Yinmin Zhong and Yishi Piao and Yisong Wang and Yixiang Zhang and Yixiao Chen and Yixuan Tan and Yixuan Wei and Yiyang Ma and Yiyuan Liu and Yonglun Yang and Yongqiang Guo and Yongtong Wu and Yu Wu and YuKun Li and Yuan Cheng and Yuan Ou and Yuanfan Xu and Yuanhao Li and Yuduan Wang and Yuehan Yang and Yuer Xu and Yuhan Wu and Yuhao Meng and Yuheng Zou and Yukun Zha and Yunfan Xiong and Yupeng Chen and Yuping Lin and Yuqian Cao and Yuqian Wang and Yushun Zhang and Yuting Yan and Yutong Lin and Yuxian Gu and Yuxiang Luo and Yuxiang You and Yuxuan Liu and Yuxuan Zhou and Yuyang Zhou and Yuzhen Huang and ZF Wu and Zehao Wang and Zehua Zhao and Zehui Ren and Zekai Zhang and Zhangli Sha and Zhe Fu and Zhe Ju and Zhean Xu and Zhenda Xie and Zhengyan Zhang and Zheren Gao and Zhewen Hao and Zhibin Gou and Zhicheng Ma and Zhigang Yan and Zhihong Shao and Zhixian Huang and Zhixuan Chen and Zhiyu Wu and Zhizhou Ren and Zhongyu Wu and Zhuoshu Li and Zhuping Zhang and Zian Xu and Zihao Wang and Zihua Qu and Zihui Gu and Zijia Zhu and Zilin Li and Zipeng Zhang and Ziwei Xie and Ziyi Gao and Ziyi Wan and Zizheng Pan and Zongqing Yao},
      year={2026},
      eprint={2606.19348},
      archivePrefix={arXiv},
      primaryClass={cs.CL},
      url={https://arxiv.org/abs/2606.19348}, 
}

@misc{openai2024gpt4technicalreport,
      title={GPT-4 Technical Report}, 
      author={OpenAI and Josh Achiam and Steven Adler and Sandhini Agarwal and Lama Ahmad and Ilge Akkaya and Florencia Leoni Aleman and Diogo Almeida and Janko Altenschmidt and Sam Altman and Shyamal Anadkat and Red Avila and Igor Babuschkin and Suchir Balaji and Valerie Balcom and Paul Baltescu and Haiming Bao and Mohammad Bavarian and Jeff Belgum and Irwan Bello and Jake Berdine and Gabriel Bernadett-Shapiro and Christopher Berner and Lenny Bogdonoff and Oleg Boiko and Madelaine Boyd and Anna-Luisa Brakman and Greg Brockman and Tim Brooks and Miles Brundage and Kevin Button and Trevor Cai and Rosie Campbell and Andrew Cann and Brittany Carey and Chelsea Carlson and Rory Carmichael and Brooke Chan and Che Chang and Fotis Chantzis and Derek Chen and Sully Chen and Ruby Chen and Jason Chen and Mark Chen and Ben Chess and Chester Cho and Casey Chu and Hyung Won Chung and Dave Cummings and Jeremiah Currier and Yunxing Dai and Cory Decareaux and Thomas Degry and Noah Deutsch and Damien Deville and Arka Dhar and David Dohan and Steve Dowling and Sheila Dunning and Adrien Ecoffet and Atty Eleti and Tyna Eloundou and David Farhi and Liam Fedus and Niko Felix and Simón Posada Fishman and Juston Forte and Isabella Fulford and Leo Gao and Elie Georges and Christian Gibson and Vik Goel and Tarun Gogineni and Gabriel Goh and Rapha Gontijo-Lopes and Jonathan Gordon and Morgan Grafstein and Scott Gray and Ryan Greene and Joshua Gross and Shixiang Shane Gu and Yufei Guo and Chris Hallacy and Jesse Han and Jeff Harris and Yuchen He and Mike Heaton and Johannes Heidecke and Chris Hesse and Alan Hickey and Wade Hickey and Peter Hoeschele and Brandon Houghton and Kenny Hsu and Shengli Hu and Xin Hu and Joost Huizinga and Shantanu Jain and Shawn Jain and Joanne Jang and Angela Jiang and Roger Jiang and Haozhun Jin and Denny Jin and Shino Jomoto and Billie Jonn and Heewoo Jun and Tomer Kaftan and Łukasz Kaiser and Ali Kamali and Ingmar Kanitscheider and Nitish Shirish Keskar and Tabarak Khan and Logan Kilpatrick and Jong Wook Kim and Christina Kim and Yongjik Kim and Jan Hendrik Kirchner and Jamie Kiros and Matt Knight and Daniel Kokotajlo and Łukasz Kondraciuk and Andrew Kondrich and Aris Konstantinidis and Kyle Kosic and Gretchen Krueger and Vishal Kuo and Michael Lampe and Ikai Lan and Teddy Lee and Jan Leike and Jade Leung and Daniel Levy and Chak Ming Li and Rachel Lim and Molly Lin and Stephanie Lin and Mateusz Litwin and Theresa Lopez and Ryan Lowe and Patricia Lue and Anna Makanju and Kim Malfacini and Sam Manning and Todor Markov and Yaniv Markovski and Bianca Martin and Katie Mayer and Andrew Mayne and Bob McGrew and Scott Mayer McKinney and Christine McLeavey and Paul McMillan and Jake McNeil and David Medina and Aalok Mehta and Jacob Menick and Luke Metz and Andrey Mishchenko and Pamela Mishkin and Vinnie Monaco and Evan Morikawa and Daniel Mossing and Tong Mu and Mira Murati and Oleg Murk and David Mély and Ashvin Nair and Reiichiro Nakano and Rajeev Nayak and Arvind Neelakantan and Richard Ngo and Hyeonwoo Noh and Long Ouyang and Cullen O'Keefe and Jakub Pachocki and Alex Paino and Joe Palermo and Ashley Pantuliano and Giambattista Parascandolo and Joel Parish and Emy Parparita and Alex Passos and Mikhail Pavlov and Andrew Peng and Adam Perelman and Filipe de Avila Belbute Peres and Michael Petrov and Henrique Ponde de Oliveira Pinto and Michael and Pokorny and Michelle Pokrass and Vitchyr H. Pong and Tolly Powell and Alethea Power and Boris Power and Elizabeth Proehl and Raul Puri and Alec Radford and Jack Rae and Aditya Ramesh and Cameron Raymond and Francis Real and Kendra Rimbach and Carl Ross and Bob Rotsted and Henri Roussez and Nick Ryder and Mario Saltarelli and Ted Sanders and Shibani Santurkar and Girish Sastry and Heather Schmidt and David Schnurr and John Schulman and Daniel Selsam and Kyla Sheppard and Toki Sherbakov and Jessica Shieh and Sarah Shoker and Pranav Shyam and Szymon Sidor and Eric Sigler and Maddie Simens and Jordan Sitkin and Katarina Slama and Ian Sohl and Benjamin Sokolowsky and Yang Song and Natalie Staudacher and Felipe Petroski Such and Natalie Summers and Ilya Sutskever and Jie Tang and Nikolas Tezak and Madeleine B. Thompson and Phil Tillet and Amin Tootoonchian and Elizabeth Tseng and Preston Tuggle and Nick Turley and Jerry Tworek and Juan Felipe Cerón Uribe and Andrea Vallone and Arun Vijayvergiya and Chelsea Voss and Carroll Wainwright and Justin Jay Wang and Alvin Wang and Ben Wang and Jonathan Ward and Jason Wei and CJ Weinmann and Akila Welihinda and Peter Welinder and Jiayi Weng and Lilian Weng and Matt Wiethoff and Dave Willner and Clemens Winter and Samuel Wolrich and Hannah Wong and Lauren Workman and Sherwin Wu and Jeff Wu and Michael Wu and Kai Xiao and Tao Xu and Sarah Yoo and Kevin Yu and Qiming Yuan and Wojciech Zaremba and Rowan Zellers and Chong Zhang and Marvin Zhang and Shengjia Zhao and Tianhao Zheng and Juntang Zhuang and William Zhuk and Barret Zoph},
      year={2024},
      eprint={2303.08774},
      archivePrefix={arXiv},
      primaryClass={cs.CL},
      url={https://arxiv.org/abs/2303.08774}, 
}

@misc{zhao2023survey,
      title={A Survey of Large Language Models}, 
      author={Wayne Xin Zhao and Kun Zhou and Junyi Li and Tianyi Tang and Xiaolei Wang and Yupeng Hou and Yingqian Min and Beichen Zhang and Junjie Zhang and Zican Dong and Yifan Du and Chen Yang and Yushuo Chen and Zhipeng Chen and Jinhao Jiang and Ruiyang Ren and Yifan Li and Xinyu Tang and Zikang Liu and Peiyu Liu and Jian-Yun Nie and Ji-Rong Wen},
      year={2026},
      eprint={2303.18223},
      archivePrefix={arXiv},
      primaryClass={cs.CL},
      url={https://arxiv.org/abs/2303.18223}, 
}

@misc{glm5team2026glm5,
      title={GLM-5: from Vibe Coding to Agentic Engineering}, 
      author={GLM-5-Team and : and Aohan Zeng and Xin Lv and Zhenyu Hou and Zhengxiao Du and Qinkai Zheng and Bin Chen and Da Yin and Chendi Ge and Chenghua Huang and Chengxing Xie and Chenzheng Zhu and Congfeng Yin and Cunxiang Wang and Gengzheng Pan and Hao Zeng and Haoke Zhang and Haoran Wang and Huilong Chen and Jiajie Zhang and Jian Jiao and Jiaqi Guo and Jingsen Wang and Jingzhao Du and Jinzhu Wu and Kedong Wang and Lei Li and Lin Fan and Lucen Zhong and Mingdao Liu and Mingming Zhao and Pengfan Du and Qian Dong and Rui Lu and Shuang-Li and Shulin Cao and Song Liu and Ting Jiang and Xiaodong Chen and Xiaohan Zhang and Xuancheng Huang and Xuezhen Dong and Yabo Xu and Yao Wei and Yifan An and Yilin Niu and Yitong Zhu and Yuanhao Wen and Yukuo Cen and Yushi Bai and Zhongpei Qiao and Zihan Wang and Zikang Wang and Zilin Zhu and Ziqiang Liu and Zixuan Li and Bojie Wang and Bosi Wen and Can Huang and Changpeng Cai and Chao Yu and Chen Li and Chengwei Hu and Chenhui Zhang and Dan Zhang and Daoyan Lin and Dayong Yang and Di Wang and Ding Ai and Erle Zhu and Fangzhou Yi and Feiyu Chen and Guohong Wen and Hailong Sun and Haisha Zhao and Haiyi Hu and Hanchen Zhang and Hanrui Liu and Hanyu Zhang and Hao Peng and Hao Tai and Haobo Zhang and He Liu and Hongwei Wang and Hongxi Yan and Hongyu Ge and Huan Liu and Huanpeng Chu and Jia'ni Zhao and Jiachen Wang and Jiajing Zhao and Jiamin Ren and Jiapeng Wang and Jiaxin Zhang and Jiayi Gui and Jiayue Zhao and Jijie Li and Jing An and Jing Li and Jingwei Yuan and Jinhua Du and Jinxin Liu and Junkai Zhi and Junwen Duan and Kaiyue Zhou and Kangjian Wei and Ke Wang and Keyun Luo and Laiqiang Zhang and Leigang Sha and Liang Xu and Lindong Wu and Lintao Ding and Lu Chen and Minghao Li and Nianyi Lin and Pan Ta and Qiang Zou and Rongjun Song and Ruiqi Yang and Shangqing Tu and Shangtong Yang and Shaoxiang Wu and Shengyan Zhang and Shijie Li and Shuang Li and Shuyi Fan and Wei Qin and Wei Tian and Weining Zhang and Wenbo Yu and Wenjie Liang and Xiang Kuang and Xiangmeng Cheng and Xiangyang Li and Xiaoquan Yan and Xiaowei Hu and Xiaoying Ling and Xing Fan and Xingye Xia and Xinyuan Zhang and Xinze Zhang and Xirui Pan and Xu Zou and Xunkai Zhang and Yadi Liu and Yandong Wu and Yanfu Li and Yidong Wang and Yifan Zhu and Yijun Tan and Yilin Zhou and Yiming Pan and Ying Zhang and Yinpei Su and Yipeng Geng and Yong Yan and Yonglin Tan and Yuean Bi and Yuhan Shen and Yuhao Yang and Yujiang Li and Yunan Liu and Yunqing Wang and Yuntao Li and Yurong Wu and Yutao Zhang and Yuxi Duan and Yuxuan Zhang and Zezhen Liu and Zhengtao Jiang and Zhenhe Yan and Zheyu Zhang and Zhixiang Wei and Zhuo Chen and Zhuoer Feng and Zijun Yao and Ziwei Chai and Ziyuan Wang and Zuzhou Zhang and Bin Xu and Minlie Huang and Hongning Wang and Juanzi Li and Yuxiao Dong and Jie Tang},
      year={2026},
      eprint={2602.15763},
      archivePrefix={arXiv},
      primaryClass={cs.LG},
      url={https://arxiv.org/abs/2602.15763}, 
}

@misc{yu2026factualknowledgebenchmarkinglearning,
      title={Beyond Factual Knowledge: Benchmarking and Learning Step-Level Procedural Rule Reasoning in Large Language Models}, 
      author={Bohan Yu and Pengfei Cao and Chen Han and Chenxi Zhou and Zhiheng Zhang and Zhiyang Xie and Wenhao Teng and Xiangwen Liao and Jun Zhao and Kang Liu},
      year={2026},
      eprint={2608.22753},
      archivePrefix={arXiv},
      primaryClass={cs.CL},
      url={https://arxiv.org/abs/2608.22753}, 
}

@misc{zhao2024expelllmagentsexperiential,
      title={ExpeL: LLM Agents Are Experiential Learners}, 
      author={Andrew Zhao and Daniel Huang and Quentin Xu and Matthieu Lin and Yong-Jin Liu and Gao Huang},
      year={2024},
      eprint={2308.10144},
      archivePrefix={arXiv},
      primaryClass={cs.LG},
      url={https://arxiv.org/abs/2308.10144}, 
}

@misc{ouyang2026reasoningbankscalingagentselfevolving,
      title={ReasoningBank: Scaling Agent Self-Evolving with Reasoning Memory}, 
      author={Siru Ouyang and Jun Yan and I-Hung Hsu and Yanfei Chen and Ke Jiang and Zifeng Wang and Rujun Han and Long T. Le and Samira Daruki and Xiangru Tang and Vishy Tirumalashetty and George Lee and Mahsan Rofouei and Hangfei Lin and Jiawei Han and Chen-Yu Lee and Tomas Pfister},
      year={2026},
      eprint={2509.25140},
      archivePrefix={arXiv},
      primaryClass={cs.AI},
      url={https://arxiv.org/abs/2509.25140}, 
}

@misc{xia2026skillrlevolvingagentsrecursive,
      title={SkillRL: Evolving Agents via Recursive Skill-Augmented Reinforcement Learning}, 
      author={Peng Xia and Jianwen Chen and Hanyang Wang and Jiaqi Liu and Kaide Zeng and Yu Wang and Siwei Han and Yiyang Zhou and Xujiang Zhao and Haifeng Chen and Zeyu Zheng and Cihang Xie and Huaxiu Yao},
      year={2026},
      eprint={2602.08234},
      archivePrefix={arXiv},
      primaryClass={cs.LG},
      url={https://arxiv.org/abs/2602.08234}, 
}

@misc{lin2026imitateboundaryawareskillmemory,
      title={When Not to Imitate: Boundary-Aware Skill Memory for Reliable Tool-Use LLM Agents}, 
      author={Zihan Lin and Zhenyu Chen and Jiawen Wei and Xiaohan Wang and Jie Cao and Jiajun Chai and Wei Lin and Guojun Yin and Ran He},
      year={2026},
      eprint={2608.22339},
      archivePrefix={arXiv},
      primaryClass={cs.CL},
      url={https://arxiv.org/abs/2608.22339}, 
}

@misc{ye2026onpolicycontextdistillationlanguage,
      title={On-Policy Context Distillation for Language Models}, 
      author={Tianzhu Ye and Li Dong and Xun Wu and Shaohan Huang and Furu Wei},
      year={2026},
      eprint={2602.12275},
      archivePrefix={arXiv},
      primaryClass={cs.CL},
      url={https://arxiv.org/abs/2602.12275}, 
}

@misc{zhang2026latentonpolicyselfdistillation,
      title={Latent On-Policy Self-Distillation}, 
      author={Guibin Zhang and Jiayang Lyu and Ran Sun and Xinlei Yu and Haoyu Zhao and Qibing Ren and Shuicheng Yan},
      year={2026},
      eprint={2608.13040},
      archivePrefix={arXiv},
      primaryClass={cs.LG},
      url={https://arxiv.org/abs/2608.13040}, 
}

@misc{yang2026opidonpolicyskilldistillation,
      title={OPID: On-Policy Skill Distillation for Agentic Reinforcement Learning}, 
      author={Shuo Yang and Jinyang Wu and Zhengxi Lu and Yuhao Shen and Fan Zhang and Lang Feng and Shuai Zhang and Haoran Luo and Zheng Lian and Zhengqi Wen and Jianhua Tao},
      year={2026},
      eprint={2606.26790},
      archivePrefix={arXiv},
      primaryClass={cs.CL},
      url={https://arxiv.org/abs/2606.26790}, 
}

@misc{shridhar2021alfworldaligningtextembodied,
      title={ALFWorld: Aligning Text and Embodied Environments for Interactive Learning}, 
      author={Mohit Shridhar and Xingdi Yuan and Marc-Alexandre Côté and Yonatan Bisk and Adam Trischler and Matthew Hausknecht},
      year={2021},
      eprint={2010.03768},
      archivePrefix={arXiv},
      primaryClass={cs.CL},
      url={https://arxiv.org/abs/2010.03768}, 
}

@misc{wang2022scienceworldagentsmarter5th,
      title={ScienceWorld: Is your Agent Smarter than a 5th Grader?}, 
      author={Ruoyao Wang and Peter Jansen and Marc-Alexandre Côté and Prithviraj Ammanabrolu},
      year={2022},
      eprint={2203.07540},
      archivePrefix={arXiv},
      primaryClass={cs.CL},
      url={https://arxiv.org/abs/2203.07540}, 
}

@misc{yao2023webshopscalablerealworldweb,
      title={WebShop: Towards Scalable Real-World Web Interaction with Grounded Language Agents}, 
      author={Shunyu Yao and Howard Chen and John Yang and Karthik Narasimhan},
      year={2023},
      eprint={2207.01206},
      archivePrefix={arXiv},
      primaryClass={cs.CL},
      url={https://arxiv.org/abs/2207.01206}, 
}

@misc{guo2025stabletoolbenchstablelargescalebenchmarking,
      title={StableToolBench: Towards Stable Large-Scale Benchmarking on Tool Learning of Large Language Models}, 
      author={Zhicheng Guo and Sijie Cheng and Hao Wang and Shihao Liang and Yujia Qin and Peng Li and Zhiyuan Liu and Maosong Sun and Yang Liu},
      year={2025},
      eprint={2403.07714},
      archivePrefix={arXiv},
      primaryClass={cs.CL},
      url={https://arxiv.org/abs/2403.07714}, 
}

@misc{qwen2026qwen35,
  title  = {{Qwen3.5}: Towards Native Multimodal Agents},
  author = {{Qwen Team}},
  year   = {2026},
  month  = {February},
  url    = {https://qwen.ai/blog?id=qwen3.5}
}

@inproceedings{
yao2023react,
title={ReAct: Synergizing Reasoning and Acting in Language Models},
author={Shunyu Yao and Jeffrey Zhao and Dian Yu and Nan Du and Izhak Shafran and Karthik R Narasimhan and Yuan Cao},
booktitle={The Eleventh International Conference on Learning Representations },
year={2023},
url={https://openreview.net/forum?id=WE_vluYUL-X}
}

@misc{grpo,
      title={DeepSeekMath: Pushing the Limits of Mathematical Reasoning in Open Language Models}, 
      author={Zhihong Shao and Peiyi Wang and Qihao Zhu and Runxin Xu and Junxiao Song and Xiao Bi and Haowei Zhang and Mingchuan Zhang and Y. K. Li and Y. Wu and Daya Guo},
      year={2024},
      eprint={2402.03300},
      archivePrefix={arXiv},
      primaryClass={cs.CL},
      url={https://arxiv.org/abs/2402.03300}, 
}

@inproceedings{
opsd,
title={Self-Distilled Reasoner: On-Policy Self-Distillation for Large Language Models},
author={Siyan Zhao and Zhihui Xie and Mengchen Liu and Jing Huang and Guan Pang and Feiyu Chen and Aditya Grover},
booktitle={Forty-third International Conference on Machine Learning},
year={2026},
url={https://openreview.net/forum?id=Jpxfof0EaS}
}

@inproceedings{
sdpo,
title={Reinforcement Learning via Self-Distillation},
author={Jonas H{\"u}botter and Frederike L{\"u}beck and Lejs Deen Behric and Anton Baumann and Marco Bagatella and Daniel Marta and Ido Hakimi and Idan Shenfeld and Thomas Kleine Buening and Carlos Guestrin and Andreas Krause},
booktitle={Forty-third International Conference on Machine Learning},
year={2026},
url={https://openreview.net/forum?id=QkfkxyRizZ}
}

@misc{skill_sd,
      title={Skill-SD: Skill-Conditioned Self-Distillation for Multi-turn LLM Agents}, 
      author={Hao Wang and Guozhi Wang and Han Xiao and Yufeng Zhou and Yue Pan and Jichao Wang and Ke Xu and Yafei Wen and Xiaohu Ruan and Xiaoxin Chen and Honggang Qi},
      year={2026},
      eprint={2604.10674},
      archivePrefix={arXiv},
      primaryClass={cs.LG},
      url={https://arxiv.org/abs/2604.10674}, 
}

@misc{sdar,
      title={Self-Distilled Agentic Reinforcement Learning}, 
      author={Zhengxi Lu and Zhiyuan Yao and Zhuowen Han and Zi-Han Wang and Jinyang Wu and Qi Gu and Xunliang Cai and Weiming Lu and Jun Xiao and Yueting Zhuang and Yongliang Shen},
      year={2026},
      eprint={2605.15155},
      archivePrefix={arXiv},
      primaryClass={cs.LG},
      url={https://arxiv.org/abs/2605.15155}, 
}

@inproceedings{song2024trial,
    title = "Trial and Error: Exploration-Based Trajectory Optimization of {LLM} Agents",
    author = "Song, Yifan  and
      Yin, Da  and
      Yue, Xiang  and
      Huang, Jie  and
      Li, Sujian  and
      Lin, Bill Yuchen",
    editor = "Ku, Lun-Wei  and
      Martins, Andre  and
      Srikumar, Vivek",
    booktitle = "Proceedings of the 62nd Annual Meeting of the Association for Computational Linguistics (Volume 1: Long Papers)",
    month = aug,
    year = "2024",
    address = "Bangkok, Thailand",
    publisher = "Association for Computational Linguistics",
    url = "https://aclanthology.org/2024.acl-long.409/",
    doi = "10.18653/v1/2024.acl-long.409",
    pages = "7584--7600"
}

@misc{openai2026gpt55,
  title  = {{GPT-5.5} System Card},
  author = {{OpenAI}},
  year   = {2026},
  month  = apr,
  url    = {https://openai.com/index/gpt-5-5-system-card/}
}

@inproceedings{qiao2024wkm,
 author = {Qiao, Shuofei and Fang, Runnan and Zhang, Ningyu and Zhu, Yuqi and Chen, Xiang and Deng, Shumin and Jiang, Yong and Xie, Pengjun and Huang, Fei and Chen, Huajun},
 booktitle = {Advances in Neural Information Processing Systems},
 doi = {10.52202/079017-3646},
 editor = {A. Globerson and L. Mackey and D. Belgrave and A. Fan and U. Paquet and J. Tomczak and C. Zhang},
 pages = {114843--114871},
 publisher = {Curran Associates, Inc.},
 title = {Agent Planning with World Knowledge Model},
 url = {https://proceedings.neurips.cc/paper_files/paper/2024/file/d032263772946dd5026e7f3cd22bce5b-Paper-Conference.pdf},
 volume = {37},
 year = {2024}
}

@inproceedings{kim2025coex,
    title = "{C}o{E}x {--} Co-evolving World-model and Exploration",
    author = "Kim, Minsoo  and
      Hwang, Seung-won",
    editor = "Christodoulopoulos, Christos  and
      Chakraborty, Tanmoy  and
      Rose, Carolyn  and
      Peng, Violet",
    booktitle = "Findings of the Association for Computational Linguistics: EMNLP 2025",
    month = nov,
    year = "2025",
    address = "Suzhou, China",
    publisher = "Association for Computational Linguistics",
    url = "https://aclanthology.org/2025.findings-emnlp.1179/",
    doi = "10.18653/v1/2025.findings-emnlp.1179",
    pages = "21629--21651",
    ISBN = "979-8-89176-335-7"
}

@misc{liu2026comap,
      title={CoMAP: Co-Evolving World Models and Agent Policies for LLM Agents}, 
      author={Youwei Liu and Jian Wang and Hanlin Wang and Wenjie Li},
      year={2026},
      eprint={2606.02372},
      archivePrefix={arXiv},
      primaryClass={cs.AI},
      url={https://arxiv.org/abs/2606.02372}, 
}

@misc{schulman2017proximalpolicyoptimizationalgorithms,
      title={Proximal Policy Optimization Algorithms}, 
      author={John Schulman and Filip Wolski and Prafulla Dhariwal and Alec Radford and Oleg Klimov},
      year={2017},
      eprint={1707.06347},
      archivePrefix={arXiv},
      primaryClass={cs.LG},
      url={https://arxiv.org/abs/1707.06347}, 
}

@inproceedings{Reflexion,
 author = {Shinn, Noah and Cassano, Federico and Gopinath, Ashwin and Narasimhan, Karthik and Yao, Shunyu},
 booktitle = {Advances in Neural Information Processing Systems},
 doi = {10.52202/075280-0377},
 editor = {A. Oh and T. Naumann and A. Globerson and K. Saenko and M. Hardt and S. Levine},
 pages = {8634--8652},
 publisher = {Curran Associates, Inc.},
 title = {Reflexion: language agents with verbal reinforcement learning},
 url = {https://proceedings.neurips.cc/paper_files/paper/2023/file/1b44b878bb782e6954cd888628510e90-Paper-Conference.pdf},
 volume = {36},
 year = {2023}
}

@InProceedings{Agent_Workflow_Memory,
  title = 	 {Agent Workflow Memory},
  author =       {Wang, Zora Zhiruo and Mao, Jiayuan and Fried, Daniel and Neubig, Graham},
  booktitle = 	 {Proceedings of the 42nd International Conference on Machine Learning},
  pages = 	 {63897--63911},
  year = 	 {2025},
  editor = 	 {Singh, Aarti and Fazel, Maryam and Hsu, Daniel and Lacoste-Julien, Simon and Berkenkamp, Felix and Maharaj, Tegan and Wagstaff, Kiri and Zhu, Jerry},
  volume = 	 {267},
  series = 	 {Proceedings of Machine Learning Research},
  month = 	 {13--19 Jul},
  publisher =    {PMLR},
  url = 	 {https://proceedings.mlr.press/v267/wang25bx.html}
}

@inproceedings{AutoGuide,
 author = {Fu, Yao and Kim, Dong-Ki and Kim, Jaekyeom and Sohn, Sungryull and Logeswaran, Lajanugen and Bae, Kyunghoon and Lee, Honglak},
 booktitle = {Advances in Neural Information Processing Systems},
 doi = {10.52202/079017-3811},
 editor = {A. Globerson and L. Mackey and D. Belgrave and A. Fan and U. Paquet and J. Tomczak and C. Zhang},
 pages = {119919--119948},
 publisher = {Curran Associates, Inc.},
 title = {AutoGuide: Automated Generation and Selection of Context-Aware Guidelines for Large Language Model Agents},
 url = {https://proceedings.neurips.cc/paper_files/paper/2024/file/d8efbb5dd415974eb095c3f06bff1f48-Paper-Conference.pdf},
 volume = {37},
 year = {2024}
}

@inproceedings{
mi2026skillpro,
title={Skill-Pro: Learning Reusable Skills from Experience via Non-Parametric {PPO} for {LLM} Agents},
author={Qirui Mi and Zhijian Ma and Mengyue Yang and Haoxuan Li and Yisen Wang and Haifeng Zhang and Jun Wang},
booktitle={Forty-third International Conference on Machine Learning},
year={2026},
url={https://openreview.net/forum?id=9kJQjx2B80}
}

@inproceedings{
agarwal2024onpolicy,
title={On-Policy Distillation of Language Models: Learning from Self-Generated Mistakes},
author={Rishabh Agarwal and Nino Vieillard and Yongchao Zhou and Piotr Stanczyk and Sabela Ramos Garea and Matthieu Geist and Olivier Bachem},
booktitle={The Twelfth International Conference on Learning Representations},
year={2024},
url={https://openreview.net/forum?id=3zKtaqxLhW}
}

@inproceedings{wang-etal-2026-reinforcement,
    title = "Reinforcement Learning for Self-Improving Agent with Skill Library",
    author = "Wang, Jiongxiao  and
      Yan, Qiaojing  and
      Wang, Yawei  and
      Tian, Yijun  and
      Mishra, Soumya Smruti  and
      Xu, Zhichao  and
      Gandhi, Megha  and
      Xu, Panpan  and
      Cheong, Lin Lee",
    editor = "Liakata, Maria  and
      Moreira, Viviane P.  and
      Zhang, Jiajun  and
      Jurgens, David",
    booktitle = "Proceedings of the 64th Annual Meeting of the {A}ssociation for {C}omputational {L}inguistics (Volume 1: Long Papers)",
    month = jul,
    year = "2026",
    address = "San Diego, California, United States",
    publisher = "Association for Computational Linguistics",
    url = "https://aclanthology.org/2026.acl-long.69/",
    doi = "10.18653/v1/2026.acl-long.69",
    pages = "1529--1550",
    ISBN = "979-8-89176-390-6"
}

@inproceedings{verl, series={EuroSys ’25},
   title={HybridFlow: A Flexible and Efficient RLHF Framework},
   url={http://dx.doi.org/10.1145/3689031.3696075},
   DOI={10.1145/3689031.3696075},
   booktitle={Proceedings of the Twentieth European Conference on Computer Systems},
   publisher={ACM},
   author={Sheng, Guangming and Zhang, Chi and Ye, Zilingfeng and Wu, Xibin and Zhang, Wang and Zhang, Ru and Peng, Yanghua and Lin, Haibin and Wu, Chuan},
   year={2025},
   month=Mar, pages={1279–1297},
   collection={EuroSys ’25} }

@misc{yu2026ruleweaverbenchmarkingrulecenteredscenario,
      title={RuleWeaver: Benchmarking Rule-Centered Scenario Reasoning for Large Language Models}, 
      author={Bohan Yu and Shi-Yang Li and Pengfei Cao and Jun Zhao and Kang Liu},
      year={2026},
      eprint={2608.26832},
      archivePrefix={arXiv},
      primaryClass={cs.CL},
      url={https://arxiv.org/abs/2608.26832}, 
}

@inproceedings{toolbench,
 author = {Qin, Yujia and Liang, Shihao and Ye, Yining and Zhu, Kunlun and Yan, Lan and Lu, Yaxi and Lin, Yankai and Cong, Xin and Tang, Xiangru and Qian, Bill and Zhao, Sihan and Hong, Lauren and Tian, Runchu and Xie, Ruobing and Zhou, Jie and Gerstein, Mark and li, dahai and Liu, Zhiyuan and Sun, Maosong},
 booktitle = {International Conference on Learning Representations},
 editor = {B. Kim and Y. Yue and S. Chaudhuri and K. Fragkiadaki and M. Khan and Y. Sun},
 pages = {9695--9717},
 title = {ToolLLM: Facilitating Large Language Models to Master 16000+ Real-world APIs},
 url = {https://proceedings.iclr.cc/paper_files/paper/2024/file/28e50ee5b72e90b50e7196fde8ea260e-Paper-Conference.pdf},
 volume = {2024},
 year = {2024}
}

@misc{kaur2026rethinkingonpolicyselfdistillationthinking,
      title={Rethinking On-Policy Self-Distillation for Thinking Models}, 
      author={Simran Kaur and Narutatsu Ri and Yinghui He and Liam Fowl and Sanjeev Arora},
      year={2026},
      eprint={2607.05184},
      archivePrefix={arXiv},
      primaryClass={cs.AI},
      url={https://arxiv.org/abs/2607.05184}, 
}

@inproceedings{sun-etal-2026-shuttle-between-symbolic-instruction,
    title = "Shuttle Between Symbolic Instructions and Neural Parameters of Large Language Models",
    author = "Sun, Wangtao  and
      Xu, Haotian  and
      Liao, Huanxuan  and
      Yu, Xuanqing  and
      Jiang, Zhongtao  and
      He, Shizhu  and
      Zhao, Jun  and
      Liu, Kang",
    editor = "Liakata, Maria  and
      Moreira, Viviane P.  and
      Zhang, Jiajun  and
      Jurgens, David",
    booktitle = "Proceedings of the 64th Annual Meeting of the {A}ssociation for {C}omputational {L}inguistics (Volume 1: Long Papers)",
    month = jul,
    year = "2026",
    address = "San Diego, California, United States",
    publisher = "Association for Computational Linguistics",
    url = "https://aclanthology.org/2026.acl-long.1269/",
    doi = "10.18653/v1/2026.acl-long.1269",
    pages = "27510--27524",
    ISBN = "979-8-89176-390-6"
}

@inproceedings{liang-etal-2026-learning-remember,
    title = "Learning How to Remember: A Meta-Cognitive Management Method for Structured and Transferable Agent Memory",
    author = "Liang, Sirui  and
      Cao, Pengfei  and
      Zhao, Jian  and
      Teng, Wenhao  and
      Liao, Xiangwen  and
      Zhao, Jun  and
      Liu, Kang",
    editor = "Liakata, Maria  and
      Moreira, Viviane P.  and
      Zhang, Jiajun  and
      Jurgens, David",
    booktitle = "Findings of the {A}ssociation for {C}omputational {L}inguistics: {ACL} 2026",
    month = jul,
    year = "2026",
    address = "San Diego, California, United States",
    publisher = "Association for Computational Linguistics",
    url = "https://aclanthology.org/2026.findings-acl.1535/",
    doi = "10.18653/v1/2026.findings-acl.1535",
    pages = "30733--30753",
    ISBN = "979-8-89176-395-1"
}

@inproceedings{guo-etal-2025-stabletoolbench-mirror-api,
    title = "{S}table{T}ool{B}ench-{M}irror{API}: Modeling Tool Environments as Mirrors of 7,000+ Real-World {API}s",
    author = "Guo, Zhicheng  and
      Cheng, Sijie  and
      Niu, Yuchen  and
      Wang, Hao  and
      Zhou, Sicheng  and
      Huang, Wenbing  and
      Liu, Yang",
    editor = "Che, Wanxiang  and
      Nabende, Joyce  and
      Shutova, Ekaterina  and
      Pilehvar, Mohammad Taher",
    booktitle = "Findings of the Association for Computational Linguistics: ACL 2025",
    month = jul,
    year = "2025",
    address = "Vienna, Austria",
    publisher = "Association for Computational Linguistics",
    url = "https://aclanthology.org/2025.findings-acl.273/",
    doi = "10.18653/v1/2025.findings-acl.273",
    pages = "5247--5270",
    ISBN = "979-8-89176-256-5"
}

@misc{yu2025srkiscalablerealtimeknowledge,
      title={SR-KI: Scalable and Real-Time Knowledge Integration into LLMs via Supervised Attention}, 
      author={Bohan Yu and Wei Huang and Kang Liu},
      year={2025},
      eprint={2511.06446},
      archivePrefix={arXiv},
      primaryClass={cs.CL},
      url={https://arxiv.org/abs/2511.06446}, 
}

@misc{yu2026premlearningpreserverefresh,
      title={PReM: Learning What to Preserve and When to Refresh for Context Compression}, 
      author={Bohan Yu and Lei Shen and Chenxi Zhou and Chen Han and Junlin Liu and Wenbo Su and Yu Cheng and Bo Zheng},
      year={2026},
      eprint={2607.14327},
      archivePrefix={arXiv},
      primaryClass={cs.CL},
      url={https://arxiv.org/abs/2607.14327}, 
}

@misc{wu2026tokmemonetokenproceduralmemory,
      title={TokMem: One-Token Procedural Memory for Large Language Models}, 
      author={Zijun Wu and Yongchang Hao and Lili Mou},
      year={2026},
      eprint={2510.00444},
      archivePrefix={arXiv},
      primaryClass={cs.CL},
      url={https://arxiv.org/abs/2510.00444}, 
}

@misc{liu2026proprietary,
      title={From Proprietary to Open-Source: Bridging the Distribution Gap via Multi-Agent Protocol Distillation in Agentic Search}, 
      author={Junlin Liu and Jiangwang Chen and Zixin Song and Shuaiyu Zhou and Chunji Lv and Hank Wu and Kailin Jiang and Jinyang Wu and Bohan Yu and Chenxi Zhou},
      year={2026},
      eprint={2607.24280},
      archivePrefix={arXiv},
      primaryClass={cs.AI},
      url={https://arxiv.org/abs/2607.24280}, 
}

@misc{wu2026contrastive,
      title={Contrastive Reinforced Policy Optimization via Privileged Self-Distillation}, 
      author={Xingjian Wu and Junlin Liu and Xingchen Liu and Xuhang Zhu and Jianing Wang and Linsen Guo and Xiaoyu Li and Xuezhi Cao and Xunliang Cai},
      year={2026},
      eprint={2607.28026},
      archivePrefix={arXiv},
      primaryClass={cs.LG},
      url={https://arxiv.org/abs/2607.28026}, 
}

@misc{lv2026pcsd,
      title={PCSD: Persistent Consistency for Self-Distillation in Agentic Reinforcement Learning}, 
      author={Chunji Lv and Yangguang Wei and Junlin Liu and Yang Gao and Ming Liu and Xinming Wang and Jinyang Wu and Guoren Wang and Changsheng Li},
      year={2026},
      eprint={2608.01837},
      archivePrefix={arXiv},
      primaryClass={cs.AI},
      url={https://arxiv.org/abs/2608.01837}, 
}

@misc{chen2026decoevo,
      title={DecoEvo: Score-Decoupled Co-Evolution of Solver and Rubric-Generator Skills in Text Space}, 
      author={Jiangwang Chen and Zixin Song and Junlin Liu and Shuaiyu Zhou and Haiyan Wu and Haihan Shi and Chenxi Zhou and Hanqing Li and Xiao Yang and Da Zhu and Guanjun Jiang and Hai Wan and Xibin Zhao},
      year={2026},
      eprint={2607.25675},
      archivePrefix={arXiv},
      primaryClass={cs.AI},
      url={https://arxiv.org/abs/2607.25675}, 
}

\clearpage
\appendix
\label{sec:appendix}

\section{Datasets}
\label{sec:dataset-details}

We evaluate AtomSD on four agent benchmarks covering household interaction, scientific experimentation, web shopping, and tool use. These benchmarks differ substantially in task structure and interaction requirements, providing a diverse testbed for procedural knowledge learning. Table~\ref{tab:dataset-statistics} summarizes their task characteristics, task diversity, and the number of task instances available for training and used for evaluation. 

\begin{table*}[h]
    \centering
    \small
    \setlength{\tabcolsep}{5pt}
    \renewcommand{\arraystretch}{1.3}
    \begin{tabular}{
        @{}
        >{\centering\arraybackslash}m{0.15\textwidth}
        m{0.40\textwidth}
        >{\centering\arraybackslash}m{0.10\textwidth}
        >{\centering\arraybackslash}m{0.08\textwidth}
        >{\centering\arraybackslash}m{0.17\textwidth}
        @{}
    }
        \toprule
        \textbf{Benchmark} &
        \multicolumn{1}{c}{\textbf{Task characteristics}} &
        \textbf{Task types} &
        \textbf{Training} &
        \textbf{Evaluation} \\
        \midrule

        ALFWorld
        & Household interaction with object search, navigation, and prerequisite-aware manipulation
        & 6 families
        & 3,553
        & 140 seen / 134 unseen \\

        ScienceWorld
        & Multi-step scientific experimentation involving transformations, measurement, circuits, and biological reasoning
        & 30 types
        & 3,592
        & 194 seen / 211 unseen \\

        WebShop
        & Interactive product search, attribute matching, option selection, and purchase
        & Open-ended
        & 10,587
        & 200 test \\

        StableToolBench
        & Multi-turn tool use involving API selection, parameterization, and execution
        & 3 groups
        & 10,129
        & 169 solvable \\

        \midrule
        \textbf{Total}
        & 
        & 
        & \textbf{27,861}
        & \textbf{1,048} \\
        \bottomrule
    \end{tabular}
    \caption{Dataset characteristics and task counts used for training and evaluation.}
    \label{tab:dataset-statistics}
\end{table*}

\paragraph{ALFWorld}
ALFWorld~\cite{shridhar2021alfworldaligningtextembodied} evaluates text-based household interaction across six task families: placing objects, placing two objects, cleaning, heating, cooling, and examining an object under a light. The benchmark requires agents to locate relevant objects, satisfy action prerequisites, and execute multi-step plans. We use 3,553 training tasks and report results on 140 seen and 134 unseen evaluation tasks.

\paragraph{ScienceWorld}
ScienceWorld~\cite{wang2022scienceworldagentsmarter5th} evaluates scientific reasoning and experimentation across 30 task types, including material transformations, electrical circuits, measurement, and biological processes. We use 3,592 training tasks and evaluate on 194 seen and 211 unseen tasks.

\paragraph{WebShop}
WebShop~\cite{yao2023webshopscalablerealworldweb} evaluates interactive shopping agents that must search a product catalog, inspect product attributes and options, and select an item satisfying a natural-language request. Its diversity arises from open-ended product constraints and shopping goals. We use 10,587 training tasks and 200 test tasks for evaluation.

\paragraph{StableToolBench}
StableToolBench~\cite{guo2025stabletoolbenchstablelargescalebenchmarking} evaluates multi-turn tool use across the G1, G2, and G3 query groups, requiring agents to select appropriate APIs, construct valid calls, and reason over tool responses. Our training pool contains 10,129 audited ToolBench~\cite{toolbench} queries, and evaluation uses the 169 solvable instances in the StableToolBench reporting subset.

\paragraph{Evaluation Subsets}
\label{sec:evaluation-subsets}
We use fixed evaluation subsets across all methods: 140 seen and 134 unseen tasks for ALFWorld, 194 seen and 211 unseen tasks for ScienceWorld, 200 test tasks for WebShop, and 169 solvable queries for StableToolBench. The ScienceWorld and WebShop subsets follow the manifests released with ETO~\cite{song2024trial}, and the same task selections are used for every method.

\section{Training and Implementation Details}
\label{sec:implementation-details}

\paragraph{Shared Setup}
We train AtomSD and all training baselines under a unified mixed-benchmark setting using verl~\cite{verl} on $8\times$ A100 80\,GB GPUs, with tasks sampled from the training pools of ALFWorld, ScienceWorld, WebShop, and StableToolBench. At each training iteration, we sample 48 tasks, with 12 drawn from each benchmark, and generate four on-policy trajectories per task. Sampling is balanced across task types for ALFWorld, ScienceWorld, and StableToolBench, while WebShop is sampled from a single task pool. Training rewards are normalized to $[0,1]$: ALFWorld and StableToolBench use binary success rewards, while ScienceWorld and WebShop retain their graded rewards after scaling by 100. We use a learning rate of $1\times10^{-6}$; other shared training hyperparameters are summarized in Table~\ref{tab:training-settings}. For AtomSD, we set the distillation coefficient to $\lambda=0.001$, the clipping range to $\epsilon=0.2$, and the EMA decay to $\alpha=0.95$, updating the teacher once per training iteration. All training baselines follow the same shared setup, with method-specific objectives, guidance sources, and teacher-update strategies summarized in Table~\ref{tab:baseline-reproduction}. For evaluation inference and Atom extraction, we set the temperature to 0.7 for Qwen3.5 models and 1.0 for API-based models. We disable thinking wherever supported; for GLM-5.3, which requires thinking, we set the reasoning effort to low.

\begin{wraptable}[21]{r}{0.56\textwidth}
    \centering
    \footnotesize
    \setlength{\tabcolsep}{4pt}
    \renewcommand{\arraystretch}{1}
    \begin{tabular}{
        @{}
        >{\raggedright\arraybackslash}m{0.43\linewidth}
        >{\raggedright\arraybackslash}m{0.49\linewidth}
        @{}
    }
        \toprule
        \multicolumn{1}{c}{\textbf{Parameter}} &
        \multicolumn{1}{c}{\textbf{Setting}} \\
        \midrule
        
        \multicolumn{2}{@{}l}{\textit{Shared training settings}} \\
        Backbones & Qwen3.5-4B / Qwen3.5-9B \\
        Tasks per iteration & 48 (12 per benchmark) \\
        Rollouts per task & 4 \\
        Optimization mini-batch & 12 tasks / 48 trajectories \\
        Optimization epochs & 1 \\
        Learning rate & $1\times10^{-6}$ \\
        Maximum context length & 49,152 tokens \\
        Rollout temperature & 1.0 \\
        Random seed & 42 \\
        
        \midrule
        \multicolumn{2}{@{}l}{\textit{AtomSD-specific settings}} \\
        KD coefficient $\lambda$ & 0.001 \\
        EMA decay $\alpha$ & 0.95 \\
        Teacher update & Every training iteration \\
        
        \bottomrule
    \end{tabular}
    \caption{Shared training hyperparameters and AtomSD-specific settings.}
    \label{tab:training-settings}
\end{wraptable}

\paragraph{Demonstration-Based Distillation}
We follow the official teacher-update settings for the demonstration-based baselines: OPSD and GRPO+OPSD use a teacher fixed at the initial backbone, while SDPO maintains an EMA teacher. Demonstrations are drawn from successful same-task rollouts in the current batch, excluding the target trajectory.

\paragraph{Skill-Based Distillation}
We reproduce Skill-SD and SDAR following the official implementation and experimental settings released with SDAR, including their respective teacher-update and distillation configurations. Both methods use a skill library as privileged teacher context, while the student receives no skill input. Skill-SD uses a distillation coefficient of 0.001, whereas SDAR uses 0.01 with its original gating mechanism; both teachers use the current student parameters at each training iteration.
For ALFWorld and WebShop, we use the skill libraries released with SDAR. For ScienceWorld and StableToolBench, where corresponding libraries are unavailable, we construct them following SDAR's official setting, using the SkillRL~\cite{xia2026skillrlevolvingagentsrecursive} procedure with training-task rollouts and GPT-5.5 as the library builder.

\paragraph{Inference-Only Baselines}
ReAct performs a single episode per task without parameter updates. Reflexion allows up to three attempts per task with task-local reflection memory, stopping early upon success. We adapt the official reflection prompt to all four benchmarks.

\begin{table*}[t]
    \centering
    \footnotesize
    \setlength{\tabcolsep}{6pt}
    \renewcommand{\arraystretch}{1}
    \begin{tabularx}{\linewidth}{
        @{}
        >{\centering\arraybackslash}m{0.13\linewidth}
        >{\centering\arraybackslash}m{0.10\linewidth}
        >{\centering\arraybackslash}m{0.09\linewidth}
        >{\centering\arraybackslash}m{0.18\linewidth}
        >{\raggedright\arraybackslash}X
        @{}
    }
        \toprule
        \textbf{Method} &
        \textbf{GRPO reward loss} &
        \textbf{Distillation weight} &
        \textbf{Teacher update} &
        \centering\arraybackslash\textbf{Distillation configuration} \\
        \midrule
        GRPO
        & Yes
        & ---
        & ---
        & Task-reward optimization only. \\

        OPSD
        & No
        & 1.0
        & Fixed
        & Sampled-token reverse-KL surrogate with successful same-task demonstrations. \\

        GRPO+OPSD
        & Yes
        & 0.001
        & Fixed
        & Same demonstration-conditioned distillation objective as OPSD. \\

        SDPO
        & No
        & 1.0
        & EMA ($\alpha=0.95$)
        & Distribution-level self-distillation over the top-$K$ tokens and the aggregated tail probability, with $K=100$. \\

        Skill-SD
        & Yes
        & 0.001
        & Full synchronization
        & Differentiable importance-weighted $k_3$ divergence. \\

        SDAR
        & Yes
        & 0.01
        & Full synchronization
        & Detached sigmoid gating with $\beta=5.0$. \\
        \bottomrule
    \end{tabularx}
    \caption{Method-specific configurations for the training baselines.}
    \label{tab:baseline-reproduction}
\end{table*}

\section{Procedural Atom Internalization and Composition Analysis}
\label{sec:atom-internalization}

We define \emph{procedural Atom internalization} as the transfer of explicit Atom guidance into the student's own policy, such that the same guidance is reflected in its behavior even when Atoms are no longer provided. We further examine whether this internalization extends to composed guidance from multiple Atoms. To examine this process, we hold the interaction history fixed and compare two sources of action-support change: the change induced by explicitly supplying Atoms to an earlier checkpoint, and the change produced by subsequent training without Atom access. We quantify their relationship using two complementary metrics: (1) \emph{maximum-token support change}, which captures the strongest local increase in support for an action, and (2) \emph{guidance-direction agreement}, which measures whether training-induced changes follow the direction specified by explicit Atom guidance. We evaluate both individual and composed Atom guidance across same-task and cross-task histories at \textit{Early} and \textit{Late} stages of training.

\paragraph{Analysis Units}
Each analysis unit consists of an interaction history, a recorded action, and an Atom pair $(\omega_\alpha,\omega_\beta)$. We obtain Atoms from successful training trajectories and use GPT-5.5 to assess their applicability to the target history and identify procedural dependencies or joint constraints relevant to the recorded action. For same-task analysis, the target history comes from a different trajectory of the same task, whereas for cross-task analysis, it comes from a task with a distinct task ID. The 4B analysis includes 175 same-task and 144 cross-task units, while the 9B analysis includes 173 and 149 units, respectively. We define \textit{Early} as training steps 0--10 and \textit{Late} as steps 50--60.

\paragraph{Conditional Action Support}
Let $u$ index an analysis unit, let $y_{uj}$ denote the token at position $j$ of its recorded action, and let $\mathcal{J}_u$ denote the positions of the scored action-content tokens. For each Atom context
$\Omega\in\{\varnothing,\{\omega_\alpha\},\{\omega_\beta\},
\{\omega_\alpha,\omega_\beta\}\}$,
we teacher-force the recorded action and compute the token-level log-probability
\begin{equation}
    \ell_{\theta,uj}(\Omega)
    =
    \log \pi_{\theta}
    \bigl(y_{uj}\mid \widetilde{h}_{uj},\Omega\bigr),
    \qquad j\in\mathcal{J}_u,
    \label{eq:internalization-support}
\end{equation}
where $\widetilde{h}_{uj}$ contains the preceding interaction history and the recorded action prefix before token $j$. We score only action-content tokens, yielding their conditional support under each Atom context.

\paragraph{Metric 1: Maximum-Token Support Change}
Let $\theta^{-}$ and $\theta^{+}$ denote the earlier and later checkpoints within an analysis interval. Without providing Atoms, we define the training-induced log-probability change for each action token as
$d_{uj}=\ell_{\theta^{+},uj}(\varnothing)-\ell_{\theta^{-},uj}(\varnothing)$.
For each analysis unit, we select the most-improved token:
\begin{equation}
    j_u^{\mathrm{update}}
    =
    \arg\max_{j\in\mathcal{J}_u} d_{uj},
    \label{eq:internalization-policy-change}
\end{equation}
with ties resolved by the earliest position. Its relative probability change is
$r_u=\exp(d_{u j_u^{\mathrm{update}}})-1$.
For a set $\mathcal{U}$ of analysis units, we report
\begin{equation}
    S_{\mathrm{max}}
    =
    \frac{100}{|\mathcal{U}|}
    \sum_{u\in\mathcal{U}} r_u.
    \label{eq:internalization-relative-support}
\end{equation}
$S_{\mathrm{max}}$ therefore measures the mean relative change in the most-improved token of each recorded action.

\begin{wraptable}[18]{r}{0.5\textwidth}
    \centering
    \scriptsize
    \setlength{\tabcolsep}{2.2pt}
    \renewcommand{\arraystretch}{1.08}
    \begin{tabular}{lllrrrrr}
        \toprule
        \textbf{Model} &
        \textbf{Stage} &
        \textbf{Hist.} &
        \textbf{$N$} &
        \textbf{$S_{\mathrm{max}}$ (\%)} &
        \textbf{$D_{\alpha}$ (\%)} &
        \textbf{$D_{\beta}$ (\%)} &
        \textbf{$D_{\alpha\beta}$ (\%)} \\
        \midrule

        \multirow{4}{*}{\textbf{4B}}
        & \multirow{2}{*}{Early}
        & Same  & 175 & 36.9 & 68.4 & 64.9 & 67.4 \\
        &
        & Cross & 144 & 16.9 & 70.8 & 72.9 & 75.0 \\
        \cmidrule(lr){2-8}
        & \multirow{2}{*}{Late}
        & Same  & 175 & 21.2 & 52.7 & 50.0 & 56.3 \\
        &
        & Cross & 144 &  5.8 & 47.1 & 46.1 & 44.2 \\

        \midrule

        \multirow{4}{*}{\textbf{9B}}
        & \multirow{2}{*}{Early}
        & Same  & 173 & 45.2 & 59.0 & 56.0 & 54.4 \\
        &
        & Cross & 149 & 25.9 & 69.9 & 68.7 & 61.9 \\
        \cmidrule(lr){2-8}
        & \multirow{2}{*}{Late}
        & Same  & 173 & 38.1 & 70.5 & 61.0 & 69.0 \\
        &
        & Cross & 149 &  8.1 & 75.8 & 70.3 & 70.5 \\

        \bottomrule
    \end{tabular}
    \caption{Procedural Atom internalization across model scales, training stages, and history settings. $N$ denotes the number of analysis units; $S_{\mathrm{max}}$ measures maximum-token support change; $D_{\alpha}$, $D_{\beta}$, and $D_{\alpha\beta}$ measure guidance-direction agreement for Atom $\omega_\alpha$, Atom $\omega_\beta$, and their composition, respectively.}
    \label{tab:internalization-flat}
\end{wraptable}

\paragraph{Metric 2: Guidance-Direction Agreement}
For each nonempty Atom context $\Omega$, we measure its explicit guidance effect at the earlier checkpoint as
\begin{equation}
    g_{uj}^{\Omega}
    =
    \ell_{\theta^{-},uj}(\Omega)
    -
    \ell_{\theta^{-},uj}(\varnothing).
    \label{eq:internalization-guidance}
\end{equation}
We then select the token with the largest guidance effect,
$j_u^{\Omega}=\arg\max_{j\in\mathcal{J}_u} g_{uj}^{\Omega}$,
independently for Atom $\omega_\alpha$, Atom $\omega_\beta$, and their combination. This selection depends only on the earlier checkpoint and is therefore independent of subsequent training changes.
At the selected token, we define directional agreement as
\begin{equation}
    c_u^{\Omega}
    =
    \mathbf{1}\!\left[
        g_{u j_u^{\Omega}}^{\Omega}
        d_{u j_u^{\Omega}} > 0
    \right].
    \label{eq:internalization-agreement}
\end{equation}
For the eligible set
$\mathcal{E}_{\Omega}
=\{u\in\mathcal{U}:g_{u j_u^{\Omega}}^{\Omega}\neq0\}$,
the agreement rate is
\begin{equation}
    D_{\Omega}
    =
    \frac{100}{|\mathcal{E}_{\Omega}|}
    \sum_{u\in\mathcal{E}_{\Omega}} c_u^{\Omega}.
\end{equation}
Thus, $D_{\Omega}$ measures how often subsequent training without Atoms changes action support in the same direction as the explicit Atom guidance.

\paragraph{Aggregation and Results}
We aggregate both metrics at the action level within each model scale, training stage, and task-matching group, such that each recorded action contributes one observation regardless of token length. Table~\ref{tab:internalization-flat} reports the aggregated results.
Maximum-token support increases across all eight groups, indicating that the selected action tokens receive stronger policy support after training. Guidance-direction agreement is also consistently observed, reaching $75.0\%$ for composed Atom guidance in \textit{Early} 4B cross-task analysis and $70.5\%$ in \textit{Late} 9B cross-task analysis. These results provide evidence that both individual and composed Atom guidance is reflected in subsequent unguided policy changes, including across different tasks.

\begin{table*}[t]
    \centering
    \small
    \setlength{\tabcolsep}{8.5pt}
    \renewcommand{\arraystretch}{2}
    \begin{tabular}{
        p{0.20\textwidth}
        p{0.22\textwidth}
        p{0.45\textwidth}
    }
        \toprule
        \textbf{Procedural Pattern} &
        \textbf{Core Dependency} &
        \textbf{Representative Case} \\
        \midrule

        \rowcolor[HTML]{FCECEF}
        \textbf{Observation-Conditioned Branching}
        &
        A downstream action depends on newly observed information.
        &
        \textbf{ScienceWorld:} AtomSD measures a substance at
        $-31\,^{\circ}\mathrm{C}$ and selects the destination specified
        for temperatures below $50\,^{\circ}\mathrm{C}$, completing the task. \\

        \rowcolor[HTML]{EAF3FB}
        \textbf{Constraint-Driven Revision}
        &
        An earlier choice must be revised after detecting a violated constraint.
        &
        \textbf{WebShop:} AtomSD rejects an over-budget product, returns to
        search, and selects a cheaper item satisfying the requested attributes. \\

        \rowcolor[HTML]{FFF5D9}
        \textbf{Subgoal-State Dependency}
        &
        A later subgoal requires the state produced by an earlier operation.
        &
        \textbf{ALFWorld:} AtomSD acquires a potato, heats it, and only then
        places the heated object in the target receptacle. \\

        \rowcolor[HTML]{EAF5EC}
        \textbf{Feedback-Guided API Recovery}
        &
        Uninformative API responses require adapting the query strategy based on returned feedback.
        &
        \textbf{StableToolBench:} After an empty response, AtomSD revises its pagination strategy, verifies the returned results, and reports nine products grounded in the API output. \\
        
        \bottomrule
    \end{tabular}
    \caption{Representative cases of recurring procedural patterns observed in evaluation trajectories.}
    \label{tab:qualitative-cases}
    \vspace{-1em}
\end{table*}

\section{Qualitative Case Studies}
\label{sec:qualitative-cases}

\subsection{Procedural Dependency Patterns}
\label{sec:procedural-dependency-patterns}
Across evaluation trajectories, we observe several recurring forms of procedural dependency that distinguish successful agent behavior beyond simply executing plausible action sequences. We highlight four representative patterns: (1) \emph{observation-conditioned branching}, (2) \emph{constraint-driven revision}, (3) \emph{subgoal-state dependency}, and (4) \emph{feedback-guided API recovery}. Table~\ref{tab:qualitative-cases} summarizes one representative case for each pattern, followed by detailed trajectory-level analysis.

\paragraph{Observation-Conditioned Branching}
This pattern arises when the appropriate downstream action depends on information revealed only through an intermediate observation. In ScienceWorld task \texttt{use-thermometer/384}, the agent must first measure an unknown substance and then select its destination according to the observed temperature. AtomSD measures $-31\,^{\circ}\mathrm{C}$ and subsequently places the substance in the box designated for temperatures below $50\,^{\circ}\mathrm{C}$, successfully completing the task. 
In contrast, unsuccessful baseline trajectories exhibit errors such as incorrect thermometer use, repeated invalid actions, and interactions with unrelated objects, failing to obtain the measurement needed for the subsequent placement decision.
This case illustrates that procedural competence requires not only acquiring relevant information, but also conditioning subsequent actions on the observed outcome.

\paragraph{Constraint-Driven Revision}
This pattern captures situations in which an initially plausible choice must be revised after newly observed information reveals a constraint violation. In WebShop task \texttt{1451}, the requested sweater must satisfy multiple constraints, including color, size, fit, and a price below \$40. AtomSD first inspects a \$100 candidate, recognizes the price mismatch, returns to search, and selects a cheaper product before configuring the requested color and size, obtaining a score of 100. 
By comparison, failed baseline trajectories exhibit incomplete constraint satisfaction, such as purchasing after selecting the color without configuring the requested size.
The case highlights the ability to revise an intermediate decision rather than persist with a candidate that no longer satisfies the task constraints.

\paragraph{Subgoal-State Dependency}
This pattern occurs when a later subgoal depends on a state established by an earlier operation. In an ALFWorld heat-and-place task, the potato must be heated before it can satisfy the final placement goal. AtomSD first acquires the potato, executes \texttt{heat potato 1 with microwave 1}, observes confirmation of the state change, and then places the heated potato in the garbage can, completing the task in eight steps. Meanwhile, failed baseline trajectories include irrelevant object manipulations that divert execution from the required acquire–heat–place sequence.
This case illustrates the need to preserve state dependencies across subgoals, ensuring that the output of one procedure satisfies the requirements of the next.

\paragraph{Feedback-Guided API Recovery}
This pattern arises when an API response is uninformative or inconsistent, requiring the agent to revise its query strategy based on returned feedback. In StableToolBench task \texttt{G1\_instruction/32617}, the user asks for all products in the \texttt{Clothing} category and their total count. AtomSD initially receives an empty result, then inspects the available categories and adjusts its requests until it retrieves nine valid products together with a reported count of nine. It subsequently verifies the result through additional pagination checks and grounds its final answer in the returned API data. 
Unsuccessful baseline trajectories instead exhibit continued querying despite obtaining nonempty results, without consolidating the retrieved information into a final answer.
This case illustrates feedback-guided recovery: when an initial tool call is uninformative, the agent uses subsequent observations to revise its interaction strategy and recover a grounded answer.

\begin{table*}[t]
    \centering
    \small
    \setlength{\tabcolsep}{4pt}
    \renewcommand{\arraystretch}{2}
    \definecolor{AtomTableOrange}{RGB}{230,97,1}
    \definecolor{AtomTableGrayBlue}{RGB}{137,153,174}
    \definecolor{AtomTableBlue}{RGB}{43,127,225}
    \definecolor{AtomTableNavy}{RGB}{43,82,130}
    \begin{tabular}{@{}>{\columncolor{AtomTableOrange!20}[0pt][\tabcolsep]}p{0.18\textwidth}>{\columncolor{AtomTableGrayBlue!18}}p{0.25\textwidth}>{\columncolor{AtomTableBlue!18}}p{0.27\textwidth}>{\columncolor{AtomTableNavy!28}[\tabcolsep][0pt]}p{0.23\textwidth}@{}}
        \toprule
        \rowcolor{white}
        \textbf{Intent} & \textbf{When} & \textbf{Do} & \textbf{Avoid} \\
        \midrule
        Collect ingredients & Required ingredients have been identified from the recipe. & Move each required ingredient into a suitable common container. & Leaving ingredients outside the container or mixing prematurely. \\
        Mix the contents & All required ingredients are in one accessible, stable container, and the recipe calls for mixing. & Mix the container and observe whether the target substance is produced. & Mixing incomplete or separated ingredients, or ignoring mixing errors. \\
        Acknowledge the product & The desired substance has been created and its location is known. & Focus on the created substance and confirm task completion. & Focusing on the wrong object or continuing without acknowledging the goal. \\
        \bottomrule
    \end{tabular}
    \caption{Paraphrases of Procedural Atoms extracted from successful training trajectories.}
    \label{tab:dough-atoms}
    \vspace{-1em}
\end{table*}

\begin{wraptable}[12]{r}{0.4\textwidth}
    \centering
    \small
    \begin{tabular}{lrl}
        \toprule
        \textbf{Method} & \textbf{Score} & \textbf{Outcome} \\
        \midrule
        GRPO & 33 & Incomplete \\
        OPSD & 25 & Incomplete \\
        GRPO+OPSD & 33 & Incomplete \\
        SDPO & 8 & Incomplete \\
        Skill-SD & 33 & Incomplete \\
        SDAR & 33 & Incomplete \\
        \midrule
        \rowcolor{ResultBest}
        \textbf{AtomSD} & \textbf{100} & Complete \\
        \bottomrule
    \end{tabular}
    \caption{Episode-level comparison on the dough task.}
    \label{tab:dough-comparison}
\end{wraptable}

\subsection{Composition of Internalized Atoms}
\label{sec:atom-composition-case}

We use a ScienceWorld chemistry case to illustrate how Procedural Atoms acquired during training can be reflected in the final model's behavior. The task requires the agent to \emph{make dough from flour and water and then focus on the resulting product}. During evaluation, the model receives only the task itself, without any Atoms or teacher guidance.

\paragraph{Atoms Learned During Training}
From earlier successful training trajectories, AtomSD extracts three relevant Atoms: (1) place the required ingredients into the same container, (2) mix them only after all ingredients are ready, and (3) focus on the target product after it has been created. Table~\ref{tab:dough-atoms} summarizes the corresponding Atoms. These Atoms are extracted from different successful training tasks, yet capture complementary procedures that can be reused and composed to solve new task instances.

\paragraph{Internalization and Composition}
In the evaluation task, AtomSD follows the same procedural structure. It places water and flour into the same glass cup, mixes them to produce dough, and then focuses on the dough to complete the task. The reappearance of these procedures without explicit Atom guidance suggests that the learned procedural knowledge has been internalized by the model, while their sequential use illustrates how multiple Atoms can be composed to solve a new interaction. As shown in Table~\ref{tab:dough-comparison}, AtomSD completes the task with a score of 100, whereas all compared baselines fail to complete it.

\begin{figure*}[t]
    \centering
    \includegraphics[width=\textwidth]{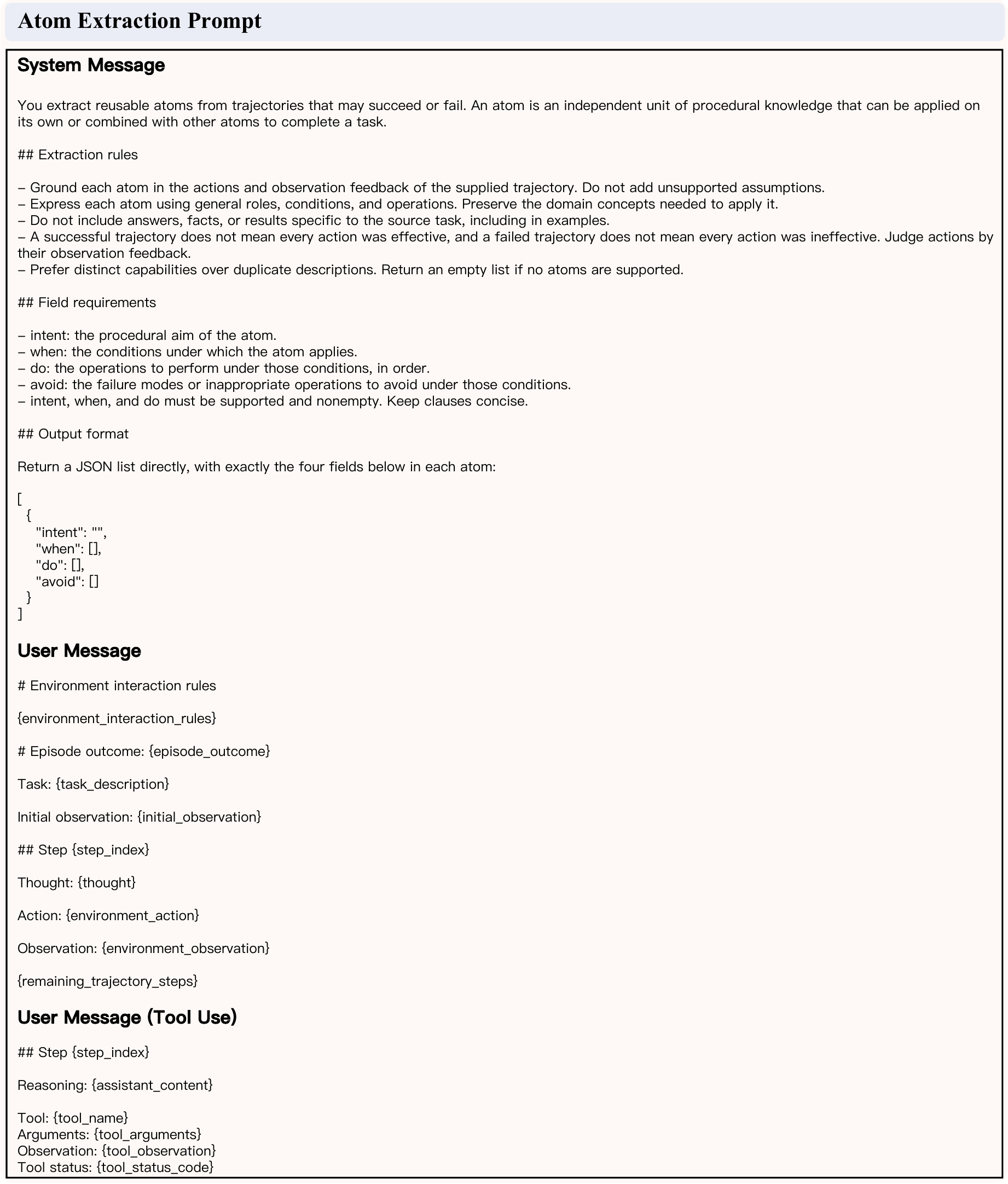}
    \caption{Prompt template for procedural Atom extraction.}
    \label{fig:prompt_extract}
    \vspace{-1em}
\end{figure*}

\begin{figure*}[t]
    \centering
    \includegraphics[width=\textwidth]{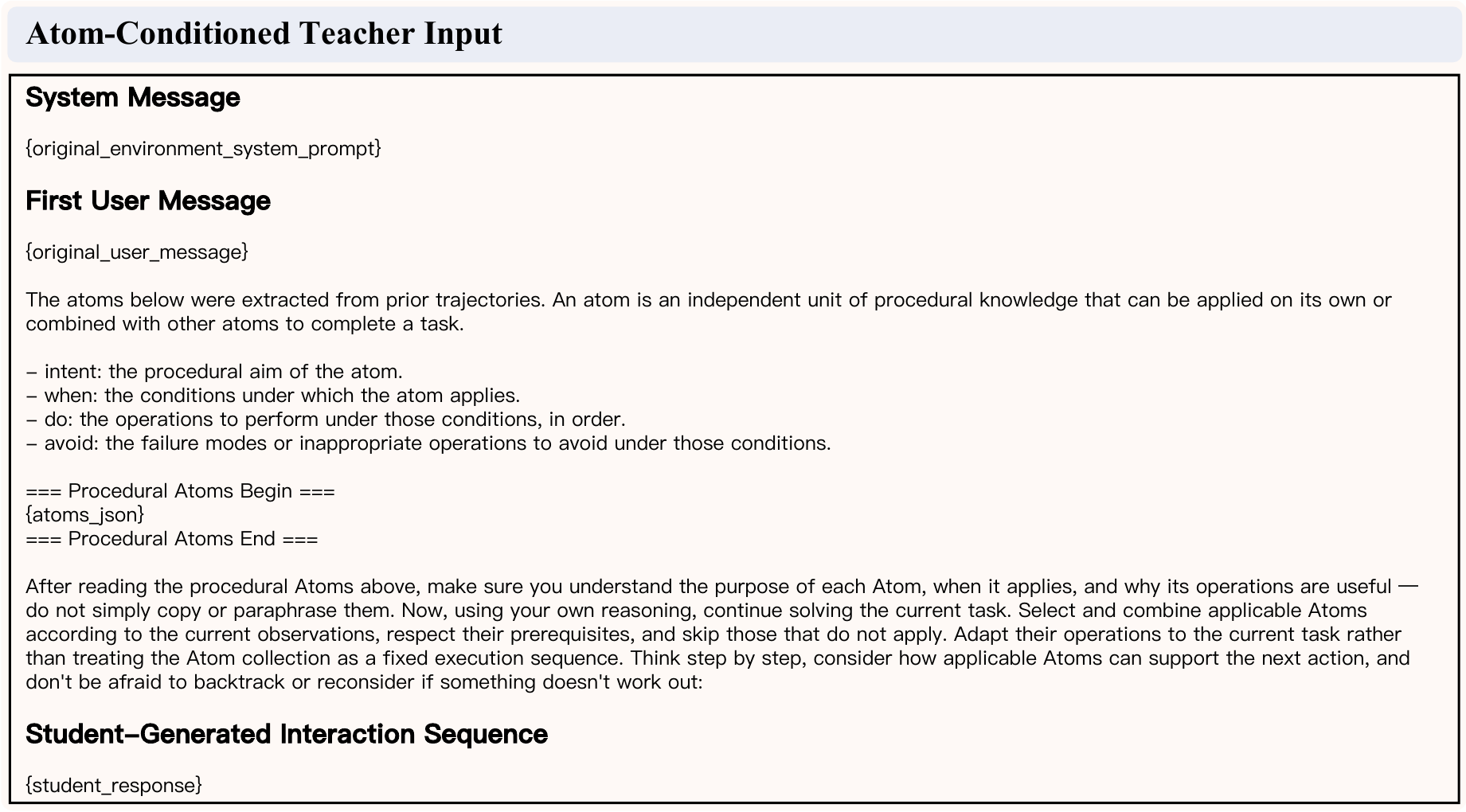}
    \caption{Atom-conditioned teacher input template.}
    \label{fig:prompt_teacher}
    \vspace{-1em}
\end{figure*}

\end{document}